\documentclass{article} 
\usepackage{iclr2027_conference,times}

\usepackage{amsmath,amsfonts,bm}

\def\eqref#1{equation~\ref{#1}}

\def\1{\bm{1}}

\DeclareMathAlphabet{\mathsfit}{\encodingdefault}{\sfdefault}{m}{sl}
\SetMathAlphabet{\mathsfit}{bold}{\encodingdefault}{\sfdefault}{bx}{n}

\usepackage{hyperref}
\usepackage{url}
\usepackage{amsmath}
\usepackage{amssymb}
\usepackage{amsthm}
\usepackage{algorithm}
\usepackage{algpseudocode}
\usepackage{caption}
\usepackage{booktabs}
\usepackage[final]{changes}
\usepackage{tikz}
\usepackage{multirow}
\usepackage{graphicx}
\usepackage{subcaption}

\newtheorem{lemma}{Lemma}

\iclrfinalcopy

\title{RAPID: Increasing Efficiency of Internal \\ Distillation via Randomized Methods}

\author{Ryan Swift \& Konstantinos Psounis \\
Department of Computer Science \\
University of Southern California \\
Los Angeles, CA 90089, USA \\
\texttt{\{ryanswif,kpsounis\}@usc.edu} \\
}

\begin{document}

\maketitle

\lhead{Preprint} 

\begin{abstract}
Knowledge distillation is a technique used to transfer the expertise of large teacher models into small, efficient student models. Recent advances have integrated the distillation of internal activations, more directly training students to learn similar representations as the teacher. One of the core challenges of internal distillation is the dimensional incompatibility between the student and teacher internal representations. Existing methods add learnable matrices to project the larger teacher activations into the smaller representation space of the student. This is a notable point of inefficiency, as learning additional parameters for the projectors increases training costs, only for them to be discarded immediately after training. We introduce RAPID (RAndomized Projection for Internal Distillation) a novel distillation methodology which eliminates the wasteful projection learning by leveraging randomized methods. RAPID utilizes an efficient transform construction based on the Johnson-Lindenstrauss lemma to compress the teacher representations into the student embedding space with strong theoretical guarantees of structural preservation. RAPID also employs a combination of exponential moving average (EMA) and homoscedastic uncertainty weighting in a novel multi-objective function designed for internal distillation. Our experiments in post-training for reasoning demonstrate that RAPID increases training efficiency by as much as 21.1\% while maintaining performance comparable to existing methods.
\end{abstract}

\section{Introduction}
\label{s:intro}
Recent open-weight large language models (LLMs), including Kimi K3 \cite{team_kimi_2026}, DeepSeek V4 Pro \cite{deepseek-ai_deepseek-v4_2026}, Qwen 3.8 Max \cite{qwen_team_qwen38-max_2026}, and GLM 5.3 \cite{glm-5-team_glm-5_2026}, have approached proprietary frontier performance across scientific reasoning, software engineering, multimodal understanding, long-context processing, and tool use. However, their substantial deployment costs make them impractical for many applications, particularly in healthcare, law, finance, cybersecurity, and autonomous vehicles, where compute, latency, and privacy constraints are stringent. Open weights enable these capabilities to be transferred into smaller, more efficient models.

Knowledge distillation transfers knowledge from a large \textit{teacher} model to a smaller \textit{student}. It has been used to pretrain models in Google's Gemma 2 and 3 families \cite{team_gemma_2024, team_gemma_2025} and Meta's Llama 3.2 family \cite{meta_llama_2024}. It is also effective for post-training \cite{guo_deepseek-r1_2025}, improving reasoning, planning, and tool-use capabilities in lightweight agentic systems.

However, text-based distillation is highly sensitive to the quality and diversity of teacher trajectories and their alignment with the student's pretraining distribution \cite{gudibande_false_2023}. \textit{Internal distillation} supplements these trajectories with teacher activations, providing richer supervision and encouraging the student to process inputs similarly to the teacher.

Existing internal-distillation methods learn projections between teacher and student representation spaces before computing losses such as mean-squared error (MSE) or KL divergence \cite{liang_less_2023, jiao_tinybert_2020, sun_patient_2019}. These methods were developed when large models contained roughly 150 million parameters, over four orders of magnitude fewer than today's approximately two-trillion-parameter frontier models. As shown in Table \ref{tab:projector-params}, at modern scales, learned projectors may require training billions of additional parameters beyond the student itself. Consequently, the cost of internal distillation remains a major barrier to further exploration and adoption.

To efficiently distill the capabilities of large modern LLMs into smaller, more efficient models, we turn to randomization. Randomization has an extensive history in accelerating numerical computation and machine learning \cite{mahoney_randomized_2011, martinsson_randomized_2021}, including specifically in dimension projection and information compression \cite{drinea_randomized_2004, beals_extensions_1984, achlioptas_database-friendly_2003}. We also draw from work in dynamic weighting and multi-objective optimization \cite{li_dynamic_2021, wang_improved_2023, kendall_multi-task_2018} to formulate a novel objective function designed for the noisy signals of internal distillation.

Our core contributions are:
\begin{itemize}
    \item In Section \ref{methodology} we introduce RAPID (\textbf{RA}ndomized \textbf{P}rojections for \textbf{I}nternal \textbf{D}istillation), a novel methodology for conducting internal distillation which eliminates the need for training projection matrices by replacing them with randomized matrices and leverages a novel dynamic multi-objective loss function
    \item In Sections \ref{results} and \ref{ss:expanded-act-set} we demonstrate that RAPID achieves task performance comparable to existing methods \deleted{for residual distillation} while improving parameter efficiency by as much as 8.1\% in the residual-only distillation scheme, and up to 21.1\% with an expanded set of distilled activations 
    \item In Section \ref{ss:responsibilities} we detail how RAPID enforces a separation of responsibilities between projection and representation learning, and Section \ref{ss:rep-align} demonstrates that RAPID produces similar or improved representational alignment with the teacher \deleted{, and demonstrate that RAPID generalizes to an expanded activation set with as much as a 21.1\% parameter efficiency improvement}
\end{itemize}

\begin{figure}[t]
\label{fig:rapid}
    \centering
    \input{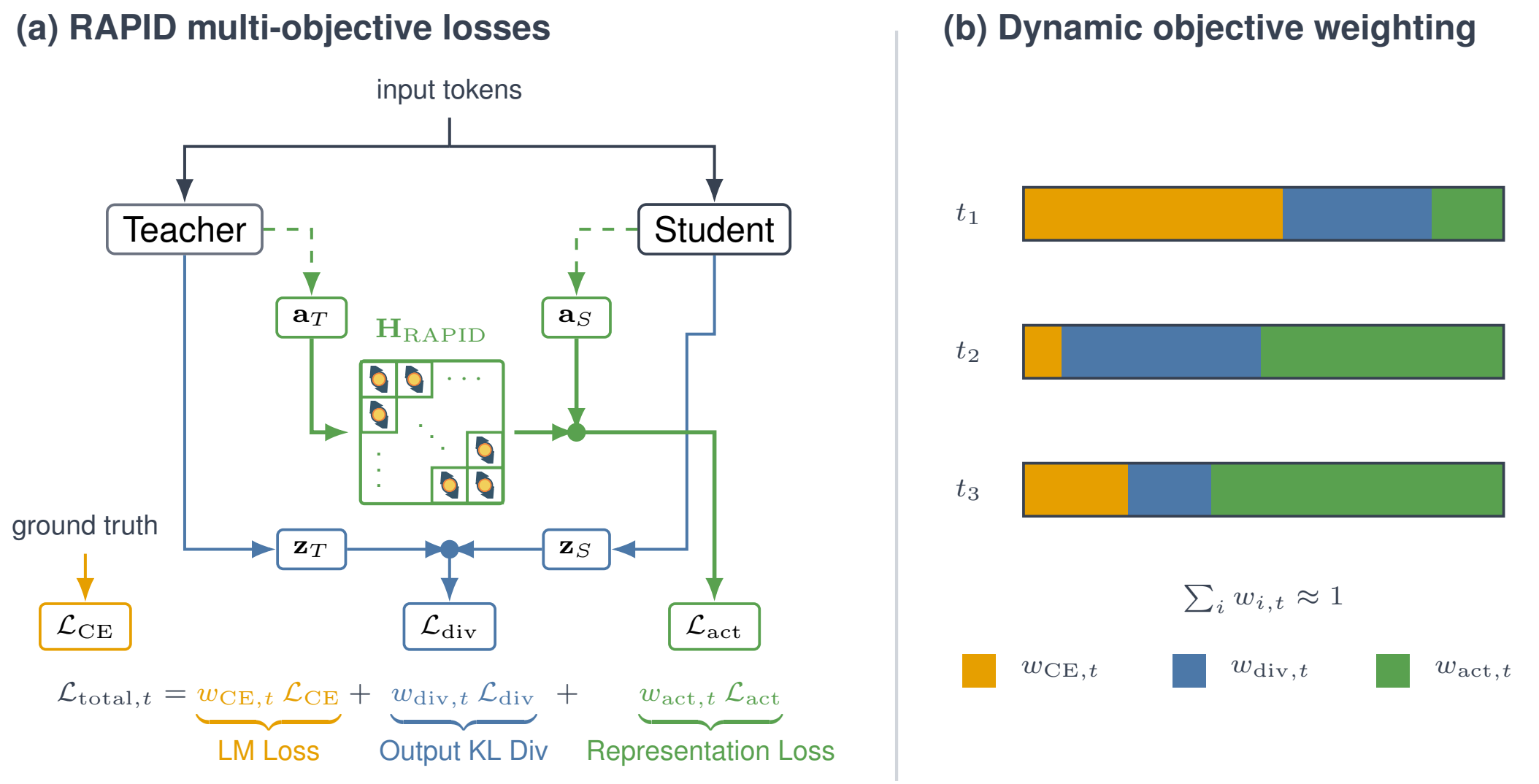}
    \caption{Overview of the RAPID internal distillation methodology. Figure \ref{fig:rapid}(a) demonstrates how teacher activations are passed through a randomized projector to produce a representation in the student's space. The multi-objective function is comprised of the language modeling loss, KL divergence on the output logits ($z_T, z_S$), and the representation loss on the internal activations ($a_T, a_S$). Figure \ref{fig:rapid}(b) shows how RAPID's dynamic weighting scheme allows the student to prioritize different forms of knowledge at different times while preventing any singlular objective from dominating or being ignored.}
    \label{fig:rapid}
\end{figure}

\section{Preliminaries}
\label{s:prelims}
\subsection{Knowledge Distillation}
Knowledge distillation (KD) was first introduced by \citet{bucilua_model_2006} and \citet{hinton_distilling_2015} was the landmark work that adapted this approach to neural networks. Recent advancements have built upon these foundational works to distill post-training behaviors such as reasoning and instruction-following using synthetic data traces \cite{guha_openthoughts_2025, guo_deepseek-r1_2025, wang_self-instruct_2023} and to distill internal activations in addition to output logits \cite{liang_less_2023, jiao_tinybert_2020, sun_patient_2019}. 

Logit and internal-activation distillation require alignment between teacher and student tokenizers. Because LLMs are sensitive to tokenization and cross-family vocabularies are difficult to align, these methods are typically restricted to models within the same family. Internal distillation must also reconcile differences in hidden dimensionality,\footnote{We use \textit{representation space}, \textit{embedding space}, \textit{internal dimension}, and \textit{hidden dimension} interchangeably.} since losses such as mean-squared error and cosine similarity require equal-dimensional vectors. Teacher and student activations must therefore be projected into a shared representation space.

\subsection{Randomized Methods for Efficient Computation}
Randomization accelerates numerical computation and machine learning by replacing expensive exact operations with efficient approximations whose errors are probabilistically controlled \cite{mahoney_randomized_2011, martinsson_randomized_2021}. One of the foundational works in randomization for informtion compression is \citet{beals_extensions_1984} which introduced the Johnson-Lindenstrauss lemma (restated here in Lemma \ref{jl-lemma}). Simply put, the JL lemma states that any set of \(n\) points in high-dimensional Euclidean space can be embedded into a smaller dimensional space while approximately preserving all pairwise distances.

\subsection{Mechanistic Interpretability}
Mechanistic interpretability seeks to develop human-interpretable accounts of how LLMs store, recognize, and process information. Although individual neurons and model components are difficult to interpret in isolation \cite{elhage_toy_2022, ameisen_circuit_2025, templeton_scaling_2024}, recent work suggests that the resulting activation spaces are not unstructured. Instead, LLMs represent concepts and semantic relationships through organized geometric patterns, including identifiable directions and low-dimensional subspaces \cite{elhage_toy_2022, bricken_towards_2023, park_linear_2024, gurnee_language_2024, park_geometry_2025}.

\section{Related Work}
\label{s:related-work}
\subsection{Knowledge Distillation}
Response-trace distillation has proven effective for transferring
post-training behaviors \cite{wang_self-instruct_2023,
guo_deepseek-r1_2025} and supports both proprietary teachers and
cross-family transfer. However, its effectiveness is sensitive to the scope, quality, and diversity of the trace distribution
\cite{gudibande_false_2023}. Logit distillation instead trains students on teacher soft targets, following \citet{hinton_distilling_2015}. Google and Meta have applied this approach during pretraining for models up to 27B parameters \cite{meta_llama_2024, team_gemma_2024, team_gemma_2025}, while Alibaba used it to consolidate multiple teacher post-training stages into a single student-training stage \cite{team_qwen3_2025}.

\subsubsection{Internal Distillation}
Internal distillation goes beyond traditional output logit distillation to additionally distill internal activations into the student \cite{liang_less_2023, jiao_tinybert_2020, sun_patient_2019}. While logit distillation attempts to improve the generalization of the knowledge being distilled by training students to replicate the teacher's output distribution, internal distillation also trains students to replicate the internal processing steps which produce the output logits. The intuition behind this approach is that the teacher is able to provide step-by-step guidance to how to student arrives at the final answer, a more granular and information-rich signal than output logits alone.

TinyBERT \cite{jiao_tinybert_2020} distills the embeddings, all residual stream activations, and all attention distribution matrices. Patient KD \cite{sun_patient_2019} reduces the computational cost of internal distillation by only distilling the internal activations of the \texttt{[CLS]} token\footnote{the \texttt{[CLS]} token in BERT models which contains the full representation of the entire sequence} from a subset of layers. TED \cite{liang_less_2023} performs task-specific distillation by training filters on top of residual stream activations to reduce the redundant information transferred to the student.

\subsection{Randomized Projections for Information Compression}

The original transform constructed by Johnson and Lindenstrauss applied a suitably scaled random orthogonal projection onto a \(k\)-dimensional subspace. Later work such as \cite{achlioptas_database-friendly_2003} improved efficiency by replacing conventional Gaussian projection matrix with one of two discrete random matrices. In the dense "binary coin" construction, each entry is sampled independently as
\begin{equation}
    R_{ij} =
    \begin{cases}
        +1, & \text{with probability } \frac{1}{2},\\
        -1, & \text{with probability } \frac{1}{2}.
    \end{cases}
\end{equation}
In the sparse ternary construction, each entry is sampled independently as
\begin{equation}
    R_{ij} = \sqrt{3}
    \begin{cases}
        +1, & \text{with probability } \frac{1}{6},\\
        0,  & \text{with probability } \frac{2}{3},\\
        -1, & \text{with probability } \frac{1}{6}.
    \end{cases}
\end{equation}
Both distributions have zero mean and unit variance, allowing the full
data matrix $\mathbf{X}\in\mathbb{R}^{n\times d}$ to be projected as
\begin{equation}
    \mathbf{Z}
    =
    \frac{1}{\sqrt{k}}\mathbf{X}\mathbf{R},
    \qquad
    \mathbf{Z}\in\mathbb{R}^{n\times k},
\end{equation}
where $\mathbf{R}\in\mathbb{R}^{d\times k}$ is the random projection
matrix. The dense construction replaces general multiplications with additions and subtractions, while the sparse construction additionally eliminates approximately two-thirds of the operations because two-thirds of its entries are zero.

\subsection{Dynamic Objective Weighting}
\citet{wang_improved_2023} demonstrated that knowledge distillation can benefit from different types of knowledge at different times during training. \citet{kendall_multi-task_2018} introduced uncertainty-based weighting for multi-objective optimization in deep learning, and \citet{li_dynamic_2021} employed it in internal distillation to allow for downweighting of representational alignment when student can already mimic teacher outputs. \citet{lakkapragada_mitigating_2022} leveraged exponential moving average (EMA) to dynamically normalize loss magnitudes for stability despite noisy updates.

\section{Methodology}
\label{methodology}

\subsection{Intuition}
There are three core features of RAPID that distinguish itself from existing internal distillation methodologies. The first is that, by using a fixed projector, we strictly enforce that the full responsibility of representation learning falls solely on the student\footnote{Further details regarding the separation of projection and representation learning are in Section \ref{ss:responsibilities}}. The second is that the usage of a JL transform provides guarantees about the preservation of structure in the projection of teacher representations. Recent mechanistic interpretability research provides evidence that LLMs represent semantic concepts and their relationships through structured geometric organization in their representation spaces \cite{elhage_toy_2022, bricken_towards_2023, park_linear_2024, gurnee_language_2024, park_geometry_2025}. By preserving teacher representational geometry, RAPID targets the distillation of semantic relationships-- a key component of advanced knowledge and intelligence. The final distinct feature is that RAPID allows the student to dynamically adapt its learning to target its weaknesses while appropriately weighting the noisy signals from distilling internal representations.

\subsection{Randomized Projector}
We use the dense binary coin JL transform described in Algorithm \ref{alg:achlioptas-dense} to create the randomized projectors in RAPID. The dense formulation was selected on the basis of modern GPUs being highly optimized for dense matrix operations. Since the JL lemma applies only to data compression, we set $d$ to the dimension of the teacher representation space and $k$ to the dimension of the student representation space to construct a \textit{downward} projection.

We construct a linearly-spaced layer mapping ($\mathcal{H}$) between the layers of the teacher and student such that every layer of the student is mapped to a layer in the teacher. For each activation being distilled, we independently initialize a separate projection matrix. For example, in our Gemma 3 experiments in Section \ref{experiments} where we distill only the residual activations, we create (18 layers) $\times$ (1 activation per layer) = 18 projectors. Our Section \ref{ss:expanded-act-set} experiments with the expanded activation set construct (18 layers) $\times$ (3 activations per layer) + (1 projector for embeddings) = 55 projectors.

\subsection{Multi-Objective Loss}
\label{ss:multi-obj-loss}
We draw from recent work in knowledge distillation \cite{li_dynamic_2021, wang_improved_2023} and utilize a dynamic objective function which combines learned uncertainty weighting \cite{kendall_multi-task_2018} with exponential moving average (EMA) progress balancing. Internal distillation is a multi-objective optimization problem, and using a dynamic objective function allows the model to prioritize different components of the objective at different points throughout training. 

At training step $t$, given an input sequence $x$, raw losses are computed between the frozen teacher $\mathcal{M}_T$ and the student $\mathcal{M}_S$ across three levels: language modeling task loss on next-token labels, output distribution divergence, and internal activation matching error.

\paragraph{Task Loss}
The student is directly trained on target sequence labels $y$ using cross-entropy loss-- the standard language modeling loss-- over student output logits $\mathbf{z}_S$:
\begin{equation}
    \mathcal{L}_{\text{CE}} = \text{CE}(\mathbf{z}_S, y).
\end{equation}

\paragraph{Output Distribution Distillation}
Output alignment between teacher logits $\mathbf{z}_T$ and student logits $\mathbf{z}_S$ is enforced using forward Kullback-Leibler (KL) divergence computed at temperature $\tau$:
\begin{equation}
    \mathcal{L}_{\text{div}}
    =
    \tau^2\,
    \operatorname{KL}\left(
        \exp\left[
            \log\operatorname{softmax}\left(\frac{\mathbf{z}_T}{\tau}\right)
        \right]
        \,\middle\|\,
        \exp\left[
            \log\operatorname{softmax}\left(\frac{\mathbf{z}_S}{\tau}\right)
        \right]
    \right).
\end{equation}

\paragraph{Intermediate Representation Distillation}
To align internal representations, we intercept internal activations for the teacher $\mathbf{a}_T^{(l)}$ and student $\mathbf{a}_S^{(l)}$ across matched hook layers $l \in \mathcal{H}$. Representation spaces are aligned using a linear projection head $P_\phi$. Depending on the projector, representation losses for internal activations $\mathcal{L}_{act}$ are computed as the Mean Squared Error (MSE) summed over matching layer pairs $\mathcal{H}$:
\begin{equation}
    \mathcal{L}_{act} = \sum_{l \in \mathcal{H}_a} \begin{cases}
        \text{MSE}\left(\mathbf{a}_T^{(l)}, P_\phi\left(\mathbf{a}_S^{(l)}\right)\right), & \text{if Projector} = \text{Learned} \\[6pt]
        \text{MSE}\left(P_\phi\left(\mathbf{a}_T^{(l)}\right), \mathbf{a}_S^{(l)}\right), & \text{if Projector} = \text{Randomized} \\[6pt]
    \end{cases}.
\end{equation}
Section \ref{results} shows results when distilling only the residual activations, though our implementation can handle an expanded set of activations (see Section \ref{ss:expanded-act-set}). Inactive internal activation objectives for a given experiment are assigned zero ($\mathcal{L}_{act} = 0$) and masked out.

\subsection{Hybrid Dynamic Loss Weighting}
\label{ss:dynamic-weight}

To dynamically balance optimization among each term in the objective, we employ a dynamic weighting mechanism which unifies learned homoscedastic uncertainty weighting with an exponential moving average (EMA) progress ratio.

Let $\mathcal{K}$ represent the set of objective components. At step $t$, let $\mathcal{L}_{i, t}$ denote the unweighted scalar loss calculated for component $i \in \mathcal{K}$.

\paragraph{Tracking Objective Baselines} 
To maintain historical loss scale baselines without retaining full step histories, an exponential moving average (EMA) buffer $\bar{\mathcal{L}}_{i, t}$ is tracked for each objective:
\begin{equation}
    \bar{\mathcal{L}}_{i, t} = \eta \bar{\mathcal{L}}_{i, t-1} + (1 - \eta) \text{sg}[\mathcal{L}_{i, t}],
\end{equation}
where $\eta \in (0, 1)$ is the EMA decay rate and $\text{sg}[\cdot]$ denotes the stop-gradient operator. We set $\eta = 0.99$.

\paragraph{Dual Signal Generation} 
The unnormalized weighting signal for objective $i$ combines two distinct mechanisms:
\begin{enumerate}
    \item \textbf{Learned Uncertainty Weighting:} Parameterized by learnable log-variance terms $s_i \in \mathbb{R}$, capturing intrinsic task noise characteristics:
    \begin{equation}
        w_i^{\text{unc}} = \exp(-s_i).
    \end{equation}
    
    \item \textbf{EMA Progress Weighting:} Measures transient loss spikes relative to historical baseline performance, scaled by hyperparameter $\alpha \ge 0$:
    \begin{equation}
        w_{i, t}^{\text{prog}} = \left( \frac{\text{sg}[\mathcal{L}_{i, t}]}{\bar{\mathcal{L}}_{i, t} + \epsilon} \right)^{\alpha},
    \end{equation}
    where $\epsilon > 0$ provides numerical stability. We set $\alpha = 0.5$ and $\epsilon = 1\times10^{-8}$.
\end{enumerate}

\paragraph{Combination, Masking, and Bounding} 
The combined raw weight for objective $i$ is calculated as the product of both signals, masked by indicator $m_i = \mathbb{I}(i \in \mathcal{A})$:
\begin{equation}
    \tilde{w}_{i, t} = m_i \cdot \left( w_i^{\text{unc}} \cdot w_{i, t}^{\text{prog}} \right).
\end{equation}
These raw weights are normalized across active objectives to project onto a probability simplex:
\begin{equation}
    \hat{w}_{i, t} = \frac{\tilde{w}_{i, t}}{\sum_{j \in \mathcal{A}} \tilde{w}_{j, t} + \epsilon}.
\end{equation}
To prevent gradient starvation or single-objective dominance, normalized weights are clamped within lower and upper bounds $[\omega_{\min}, \omega_{\max}]$ for active tasks and re-normalized:
\begin{equation}
\label{eq:weight-clip}
    w_{i, t}' = \begin{cases} 
        \text{clip}\left(\hat{w}_{i, t}, \omega_{\min}, \omega_{\max}\right), & \text{if } i \in \mathcal{A} \\ 
        0, & \text{otherwise} 
    \end{cases}, \qquad
    w_{i, t} = \frac{w_{i, t}'}{\sum_{j \in \mathcal{A}} w_{j, t}' + \epsilon}.
\end{equation}
We use $\omega_{min} = 0.05$, $\omega_{max} = 0.80$.

\paragraph{Total Objective Function} 
The final scalar loss minimized at step $t$ combines dynamically scaled raw objectives with a linear log-variance regularization term over active tasks:
\begin{equation}
    \mathcal{L}_{\text{total}} = \sum_{i \in \mathcal{A}} w_{i, t} \mathcal{L}_{i, t} + \sum_{i \in \mathcal{A}} s_i.
\end{equation}

\paragraph{Optimization Mechanics} All operations computing intermediate weights $w_{i, t}$ from $\mathcal{L}_{i, t}$ and $s_i$ are evaluated without gradient tracking. As a result, $\partial \mathcal{L}_{\text{total}} / \partial s_i = 1$ for all active heads, preventing representation backpropagation from corrupting $s_i$ and avoiding scale-invariance degeneracies. Further details regarding this are provided in Appendix \ref{a:stop_gradient_proof}.

\section{Experiments}
\label{experiments}

\subsection{Comparisons}
We compare RAPID with internal distillation using learned linear projectors, the most common approach in prior work \cite{xu_survey_2024, zhu_survey_2024}. Following existing methods \cite{jiao_tinybert_2020, sanh_distilbert_2020, liang_less_2023, sun_patient_2019}, each projector maps a student activation upward into the teacher's representation space. We use separate projectors for each distilled activation and equally weight the ground-truth, output-distribution, and internal-activation losses.

\subsection{Models}
\label{ss:models}
We demonstrate RAPID using distillation pairs from the Gemma 3 \cite{team_gemma_2025} and Qwen 2.5 \cite{qwen_qwen25_2025} model families. We exclusively perform in-family distillation, distilling Gemma 3 4B IT into Gemma 3 270M PT\footnote{Gemma 3 270M actually has 436M parameters, but the embedding and unembedding weights are tied}, and Qwen 2.5 3B Instruct into Qwen 2.5 0.5B. These approximate parameter compression ratios of 9.2:1 for Gemma 3 and 6:1 for Qwen 2.5 follow the existing distillation literature range of 1.6:1 to 10:1 \cite{jiao_tinybert_2020, sanh_distilbert_2020, meta_llama_2024, gu_minillm_2024} with absolute parameter counts scaled to modern LLM sizes.

We selected the student initialization point to be the base pretrained models for the cleanest starting point for our post-training. Each model is trained with three separate seeds \{3, 33, 333\} and the scores reported are averaged across the seeds.

\subsection{Training}
We train on 40M tokens of reasoning data with a sequence length of 1024\footnote{1024 was the largest sequence length our hardware could accommodate}, corresponding to about 40k training samples per experiment. 
To show data generalization, we trained with 2 separate datasets: a version of SmolTalk \cite{allal_smollm2_2025} curated for reasoning and maximum 1024 sequence length per sample, and unmodified OpenThoughts-114k \cite{guha_openthoughts_2025}. Given the chat format of these data, we use the tokenizer for the instruction-tuned variants of the student models.


\subsection{Evaluation Tasks}
We evaluate reasoning task performance on four benchmarks: MMLU \cite{hendrycks_measuring_2021}, ARC-Challenge \cite{clark_think_2018}, and the Open Leaderboard V2 \cite{fourrier_open_2024} versions of BBH \cite{suzgun_challenging_2022} and GPQA \cite{rein_gpqa_2023}. These are common benchmarks used in the literature for LLMs of similar size to our student models \cite{qwen_qwen25_2025, team_qwen3_2025, team_gemma_2024, team_gemma_2025, biderman_pythia_2023, meta_llama_2024} and are designed to evaluate knowledge and reasoning capabilities. Following the conventions in the literature, we evaluate MMLU as 5-shot, ARC-Challenge as 25-shot, BBH as 3-shot, and GPQA as 0-shot.

\section{Results}
\label{results}

\subsection{Task Performance}
\label{ss:perf}

As shown in Table \ref{tab:perf}, two methods achieve nearly identical performance despite RAPID being over 6\% more efficient in the Gemma 3 experiments, and over 8\% more efficient in the Qwen 2.5 experiments. No difference exceeds 0.15 percentage points in the overall average, with largest individual delta being +1.05 percentage points for RAPID over Learned for SmolTalk on BBH. 

\begin{table*}[t]
\centering
\small
\setlength{\tabcolsep}{5pt}
\begin{tabular}{lclccccc}
\toprule
\textbf{Dataset} &
\textbf{Distillation Pair} &
\textbf{Variant} &
\textbf{ARC-C} &
\textbf{BBH} &
\textbf{GPQA} &
\textbf{MMLU} &
\textbf{Average} \\
\midrule

\multirow{4}{*}{OpenThoughts}
& \multirow{2}{*}{Gemma 3}
    & Learned
    & \textbf{0.2770}
    & 0.2804
    & 0.2438
    & \textbf{0.2475}
    & \textbf{0.2622} \\
&
&
    RAPID
    & 0.2747
    & \textbf{0.2848}
    & \textbf{0.2506}
    & 0.2381
    & 0.2621 \\
\cmidrule(lr){2-8}
&
\multirow{2}{*}{Qwen 2.5}
    & Learned
    & \textbf{0.3376}
    & \textbf{0.3018}
    & 0.2508
    & \textbf{0.2560}
    & \textbf{0.2866} \\
&
&
    RAPID
    & 0.3348
    & 0.3005
    & \textbf{0.2539}
    & 0.2520
    & 0.2853 \\

\midrule

\multirow{4}{*}{SmolTalk}
& \multirow{2}{*}{Gemma 3}
    & Learned
    & 0.2742
    & \textbf{0.2898}
    & \textbf{0.2550}
    & 0.2355
    & 0.2636 \\
&
&
    RAPID
    & \textbf{0.2796}
    & 0.2895
    & 0.2517
    & \textbf{0.2382}
    & \textbf{0.2647} \\
\cmidrule(lr){2-8}
&
\multirow{2}{*}{Qwen 2.5}
    & Learned
    & 0.3490
    & 0.3129
    & \textbf{0.2581}
    & \textbf{0.3241}
    & 0.3110 \\
&
&
    RAPID
    & \textbf{0.3549}
    & \textbf{0.3234}
    & 0.2503
    & 0.3156
    & \textbf{0.3111} \\

\bottomrule
\end{tabular}
\caption{Performance comparison between learned projectors and RAPID across 2
datasets and 2 distillation pairs in the residual distillation scheme. Reported scores are averages across 3 seeds.}
\label{tab:perf}
\end{table*}

\subsection{Efficiency}
RAPID improves training efficiency by eliminating learned projectors. Table~\ref{tab:projector-params} reports their additional parameter costs across internal distillation configurations, where \textbf{All} includes residual, MLP-output, attention-output, and embedding activations. We calculate RAPID's efficiency gain as
\begin{equation}
    \frac{P_{\mathrm{added}}}
    {P_{\mathrm{model}} + P_{\mathrm{added}}}
    \times 100\%.
\end{equation}
Learned projectors increase trainable parameters by approximately 10\% for residual-only distillation and 30\% for \textbf{All} activations---up to 2B additional parameters for a 7B student. RAPID reduces trainable parameters by 8.1\% in our Qwen experiments (Section~\ref{ss:perf}) and 21.1\% when distilling \textbf{All} activations (Section~\ref{ss:expanded-act-set}).

\begin{table}[t]
\centering
\small
\begin{tabular}{llcccccc}
\toprule
\textbf{Teacher} &
\textbf{Student} &
\multicolumn{2}{c}{\textbf{Added Params}} &
\multicolumn{2}{c}{\textbf{Added Params (\%)}} &
\multicolumn{2}{c}{\textbf{RAPID Efficiency}} \\
\cmidrule(lr){3-4}
\cmidrule(lr){5-6}
\cmidrule(lr){7-8}
& &
\textbf{Residual} &
\textbf{All} &
\textbf{Residual} &
\textbf{All} &
\textbf{Residual} &
\textbf{All} \\
\midrule
Gemma 3 4B   & Gemma 3 270M
    & 29.5M  & 90.1M & 6.8\%  & 20.7\% & +6.3\%  & +17.1\% \\
Qwen 2.5 3B  & Qwen 2.5 0.5B
    & 44M    & 134M  & 8.8\%  & 26.8\% & +8.1\%  & +21.1\% \\
Llama 3.1 8B & Llama 3.2 1B
    & 134.2M & 411M  & 13.4\% & 41.1\% & +11.8\% & +29.1\% \\
Olmo 3 32B   & Olmo 3 7B
    & 671.1M & 2.03B & 9.6\%  & 29.0\% & +8.8\%  & +22.5\% \\
\bottomrule
\end{tabular}
\caption{Projector parameter overhead for each teacher--student model pairing and the parameter efficiency gain from eliminating these additional parameters with RAPID. We use 436M as the parameter count for Gemma 3 270M. We provide additional model pairings which represent reasonable candidates for distillation.}
\label{tab:projector-params}
\end{table}

\section{Analysis}
\label{analysis}

\subsection{Projector--Student Separation of Responsibilities}
\label{ss:responsibilities}
Internal-activation distillation ideally assigns distinct roles to the
student and projector: the student learns the transferred teacher
representation, while the projector performs a transformation to reconcile dimensionality.
RAPID enforces this separation by applying a fixed downward projection
$\mathbf{H}_{\mathrm{fix}}\in\mathbb{R}^{d_s\times d_t}$ to the teacher:
\begin{equation}
    \mathcal{L}_{\mathrm{down}}
    =
    \frac{1}{d_s}
    \left\|
        \mathbf{a}_s
        -
        \mathbf{H}_{\mathrm{fix}}\mathbf{a}_t
    \right\|_2^2.
\end{equation}
Because the projected teacher activation is stationary, the projector
only defines the target; all adaptation required to reduce the loss
must occur in the student.

Existing methods commonly utilize a learned upward projection
$\mathbf{H}_{\mathrm{learn}}\in\mathbb{R}^{d_t\times d_s}$
\cite{liang_less_2023, jiao_tinybert_2020, sun_patient_2019}:
\begin{equation}
    \mathcal{L}_{\mathrm{up}}
    =
    \frac{1}{d_t}
    \left\|
        \mathbf{H}_{\mathrm{learn}}\mathbf{a}_s-\mathbf{a}_t
    \right\|_2^2.
\end{equation}
Learned projectors are popular for their flexibility: beyond matching dimensions, they can learn particular transforms which facilitate student learning \cite{miles_understanding_2024, chen_improved_2022}. However, jointly training the projector and student allows both to reduce the loss. In upward projections, the projector determines both the activation residual and how it reaches the student, potentially distributing alignment across the student and a projector discarded after training. We call this effect \textit{dilution}.

RAPID trades this flexibility for a strict separation of responsibilities by removing the competing adaptation path and placing representation learning entirely on the student, eliminating the possibility of dilution. Appendix~\ref{app:responsibilities} provides the corresponding gradient analysis.

\subsection{Representational Alignment}
\label{ss:rep-align}

We evaluate representational alignment between students and teachers using centered kernel alignment (CKA) and representational similarity analysis (RSA), the latter with an inner metric of cosine similarity and an outer metric of Spearman correlation. CKA measures the overall alignment between representations and RSA measures the similarity of pairwise relationships. We analyzed the students trained on SmolTalk in the residual-only distillation scheme from Section \ref{results}. Final token activations were collected from students and teachers across 3 seeds \{123, 45, 67\} for 500 samples of the train set of GSM8K \cite{cobbe_training_2021} with a 1024 sequence length. Tables and figures for these results can be found in Appendix \ref{a:rep-align}. \textbf{Global} refers to an analysis of the full set of teacher layers vs the full set of student layers, and \textbf{Mapped} only uses the set of teacher layers in our distillation mapping $\mathcal{H}$. $P_{25}$ is the 25th percentile value.

RAPID produces substantially stronger RSA values for Gemma 3 across all global and mapped metrics, while the learned projector performs slightly better on most CKA metrics. For Qwen 2.5, the learned projector consistently—but only marginally—outperforms RAPID across both CKA and RSA. Overall, the small differences suggest that RAPID achieves representational alignment comparable to learned projection without requiring training additional projection parameters. Interestingly, the Qwen 2.5 values are significantly higher than the Gemma 3 values, even though the entire Gemma 3 family was pretrained with distillation \cite{team_gemma_2025}. We leave an explanation of this to future work.

\subsection{Distilling Additional Activations}
\label{ss:expanded-act-set}
While some existing internal distillation methods distill only the residual stream activations \cite{sun_patient_2019, liang_less_2023}, other methods such as TinyBERT \cite{jiao_tinybert_2020} distill an expanded set of activations. To demonstrate that RAPID generalizes to an expanded activation set, we include an analysis which distills attention outputs, MLP outputs, and embeddings in addition to the residual activations on our Smoltalk dataset. We refer to this expanded set of activations as \textbf{All} activations.

Table \ref{tab:all-acts} contains the results for distilling \textbf{All} activations for the SmolTalk dataset. The Gemma 3 distillation pair achieves very similar results as in the residual-only distillation scheme in Table \ref{tab:perf}, however there is a small performance degradation for Qwen 2.5, with RAPID performing 2.2 percentage points worse when distilling \textbf{All} activations. These Qwen 2.5 results also exhibit the largest performance gap of all of our experiments, at +1.67 percentage points for the learned projection. However, this small performance lift for learned projectors over RAPID comes at the cost of over 20\% worse efficiency.

\begin{table}[t]
\label{tab:all-acts}
\centering
\small
\setlength{\tabcolsep}{5pt}
\begin{tabular}{lcccccc}
\toprule
\textbf{Distillation Pair} &
\textbf{Variant} &
\textbf{ARC-C} &
\textbf{BBH} &
\textbf{GPQA} &
\textbf{MMLU} &
\textbf{Average} \\
\midrule

\multirow{2}{*}{Gemma 3}
    & Learned
    & 0.2688
    & \textbf{0.2881}
    & \textbf{0.2550}
    & 0.2363
    & 0.2621 \\
    & RAPID
    & \textbf{0.2779}
    & \textbf{0.2881}
    & \textbf{0.2550}
    & \textbf{0.2386}
    & \textbf{0.2649} \\

\midrule

\multirow{2}{*}{Qwen 2.5}
    & Learned
    & \textbf{0.3461}
    & \textbf{0.3115}
    & 0.2562
    & \textbf{0.3092}
    & \textbf{0.3058} \\
    & RAPID
    & 0.3441
    & 0.3027
    & \textbf{0.2578}
    & 0.2521
    & 0.2892 \\

\bottomrule
\end{tabular}
\caption{Performance comparison between learned projection and RAPID on
SmolTalk, distilling \textbf{All} activations. }
\label{tab:all-acts}
\end{table}

\section{Limitations}
\label{limitations}
Limited compute\footnote{We ran all experiments on a single server with two RTX A6000 48GB GPUs.} constrained model size, sequence length, and teacher finetuning, preventing full distillation and evaluation on R1-style long-context reasoning. Moreover, we evaluate RAPID only during reasoning-focused supervised finetuning and do not assess its effectiveness in pretraining, mid-training, general instruction tuning, tool-use training, or as an initialization for subsequent reinforcement learning and other post-training stages. Future work should evaluate RAPID across these training phases and at larger scales, as well as optimize the dynamic-weight bounds $\omega_{\min}, \omega_{\max}$ in Equation~\ref{eq:weight-clip}.

\section{Conclusion}
\label{conclusion}

We introduced RAPID, a novel methodology for internal distillation which achieves comparable performance to existing methods with learned projectors while being 6.3\%--21.2\% more efficient in our experiments. RAPID generalizes across models families, training datasets, and activation sets. RAPID places the responsibility of representation learning on the student, resulting in comparable or better representational similarity with the teacher.

\section*{AI use statement}

In this work, we used generative AI tools for: formulating mathematical claims, providing critical ingredients for proving mathematical claims, and assistance in writing proofs in Appendices \ref{a:stop_gradient_proof} and \ref{app:responsibilities}; designing and providing feedback on the objective formulation; and implementing methods in our experimental harness. We have not used generative AI tools for helping develop theoretical models or conceptual frameworks, proposing or refining hypotheses, supporting qualitative and thematic data analysis, or interpreting results.and generating synthetic datasets, assisting with translation, and cleaning and reformatting data are not applicable to this work. 

Additionally, we used generative AI tools for suggesting experimental parameters, creating and editing code, drafting Sections \ref{ss:multi-obj-loss} and \ref{ss:dynamic-weight}, summarizing existing literature, identifying research gaps, brainstorming, sourcing/searching for information, editing Sections \ref{s:intro}, \ref{s:prelims}, and \ref{s:related-work} for brevity, identifying relevant literature, and formatting references. 

We have reviewed all AI-assisted work. We checked LLM-generated research ideas for potential plagiarism through a manual literature survey. LLM-generated code was verified and tested for correctness by the first author. LLM-generated proofs were reviewed and modified for correctness and clarity. We take responsibility for the final content of this work, including text, claims or artifacts produced with the aid of generative AI.

\section*{Reproducibility statement}

We have taken a number of steps to ensure high levels of reproducibility in our work. All experiments were run with random seeds (whose values we provided) and the PyTorch reproducible backend for NVIDIA GPUs. Algorithm \ref{alg:achlioptas-dense} contains all details needed to reproduce the randomized projectors, and Sections \ref{ss:multi-obj-loss} and \ref{ss:dynamic-weight} show detailed formulations of our multiple objectives, dynamic weight updates, and parameters used. 

\newpage

\bibliography{references}

@misc{fourrier_open_2024,
	title = {Open {LLM} {Leaderboard} v2},
	url = {https://huggingface.co/spaces/open-llm-leaderboard/open_llm_leaderboard},
	publisher = {Hugging Face},
	author = {Fourrier, Clémentine and Habib, Nathan and Lozovskaya, Alina and Szafer, Konrad and Wolf, Thomas},
	year = {2024},
}

@misc{ameisen_circuit_2025,
	title = {Circuit {Tracing}: {Revealing} {Computational} {Graphs} in {Language} {Models}},
	shorttitle = {Circuit {Tracing}},
	url = {https://transformer-circuits.pub/2025/attribution-graphs/methods.html},
	language = {en},
	urldate = {2025-12-01},
	journal = {Transformer Circuits},
	author = {Ameisen*, Authors Emmanuel and Lindsey*, Jack and Pearce*, Adam and Gurnee*, Wes and Turner*, Nicholas L. and Chen*, Brian and Citro*, Craig and Abrahams, David and Carter, Shan and Hosmer, Basil and Marcus, Jonathan and Sklar, Michael and Templeton, Adly and Bricken, Trenton and McDougall, Callum and Cunningham, Hoagy and Henighan, Thomas and Jermyn, Adam and Jones, Andy and Persic, Andrew and Qi, Zhenyi and Thompson, T. Ben and Zimmerman, Sam and Rivoire, Kelley and Conerly, Thomas and Olah, Chris and March 27, Joshua Batson*‡ Affiliations Anthropic Published},
	month = mar,
	year = {2025},
}

@misc{zhu_survey_2024,
	title = {A {Survey} on {Model} {Compression} for {Large} {Language} {Models}},
	url = {https://arxiv.org/abs/2308.07633},
	author = {Zhu, Xunyu and Li, Jian and Liu, Yong and Ma, Can and Wang, Weiping},
	year = {2024},
}

@misc{wang_self-instruct_2023,
	title = {Self-{Instruct}: {Aligning} {Language} {Models} with {Self}-{Generated} {Instructions}},
	shorttitle = {Self-{Instruct}},
	url = {http://arxiv.org/abs/2212.10560},
	doi = {10.48550/arXiv.2212.10560},
	urldate = {2026-03-22},
	publisher = {arXiv},
	author = {Wang, Yizhong and Kordi, Yeganeh and Mishra, Swaroop and Liu, Alisa and Smith, Noah A. and Khashabi, Daniel and Hajishirzi, Hannaneh},
	month = may,
	year = {2023},
	note = {arXiv:2212.10560 [cs]},
}

@article{templeton_scaling_2024,
	title = {Scaling {Monosemanticity}: {Extracting} {Interpretable} {Features} from {Claude} 3 {Sonnet}},
	url = {https://transformer-circuits.pub/2024/scaling-monosemanticity/index.html},
	journal = {Transformer Circuits Thread},
	author = {Templeton, Adly and Conerly, Tom and Marcus, Jonathan and Lindsey, Jack and Bricken, Trenton and Chen, Brian and Pearce, Adam and Citro, Craig and Ameisen, Emmanuel and Jones, Andy and Cunningham, Hoagy and Turner, Nicholas L and McDougall, Callum and MacDiarmid, Monte and Freeman, C. Daniel and Sumers, Theodore R. and Rees, Edward and Batson, Joshua and Jermyn, Adam and Carter, Shan and Olah, Chris and Henighan, Tom},
	year = {2024},
}

@inproceedings{liang_less_2023,
	address = {Honolulu, Hawaii, USA},
	series = {{ICML}'23},
	title = {Less is more: task-aware layer-wise distillation for language model compression},
	volume = {202},
	shorttitle = {Less is more},
	urldate = {2025-10-13},
	booktitle = {Proceedings of the 40th {International} {Conference} on {Machine} {Learning}},
	publisher = {JMLR.org},
	author = {Liang, Chen and Zuo, Simiao and Zhang, Qingru and He, Pengcheng and Chen, Weizhu and Zhao, Tuo},
	month = jul,
	year = {2023},
	pages = {20852--20867},
}

@article{bricken_towards_2023,
	title = {Towards {Monosemanticity}: {Decomposing} {Language} {Models} {With} {Dictionary} {Learning}},
	journal = {Transformer Circuits Thread},
	author = {Bricken, Trenton and Templeton, Adly and Batson, Joshua and Chen, Brian and Jermyn, Adam and Conerly, Tom and Turner, Nick and Anil, Cem and Denison, Carson and Askell, Amanda and Lasenby, Robert and Wu, Yifan and Kravec, Shauna and Schiefer, Nicholas and Maxwell, Tim and Joseph, Nicholas and Hatfield-Dodds, Zac and Tamkin, Alex and Nguyen, Karina and McLean, Brayden and Burke, Josiah E and Hume, Tristan and Carter, Shan and Henighan, Tom and Olah, Christopher},
	year = {2023},
}

@inproceedings{biderman_pythia_2023,
	title = {Pythia: {A} suite for analyzing large language models across training and scaling},
	booktitle = {International {Conference} on {Machine} {Learning}},
	publisher = {PMLR},
	author = {Biderman, Stella and Schoelkopf, Hailey and Anthony, Quentin Gregory and Bradley, Herbie and O’Brien, Kyle and Hallahan, Eric and Khan, Mohammad Aflah and Purohit, Shivanshu and Prashanth, USVSN Sai and Raff, Edward and {others}},
	year = {2023},
	pages = {2397--2430},
}

@misc{cobbe_training_2021,
	title = {Training {Verifiers} to {Solve} {Math} {Word} {Problems}},
	url = {https://arxiv.org/abs/2110.14168},
	author = {Cobbe, Karl and Kosaraju, Vineet and Bavarian, Mohammad and Chen, Mark and Jun, Heewoo and Kaiser, Lukasz and Plappert, Matthias and Tworek, Jerry and Hilton, Jacob and Nakano, Reiichiro and Hesse, Christopher and Schulman, John},
	year = {2021},
}

@misc{allal_smollm2_2025,
	title = {{SmolLM2}: {When} {Smol} {Goes} {Big} – {Data}-{Centric} {Training} of a {Small} {Language} {Model}},
	url = {https://arxiv.org/abs/2502.02737},
	author = {Allal, Loubna Ben and Lozhkov, Anton and Bakouch, Elie and Blázquez, Gabriel Martín and Penedo, Guilherme and Tunstall, Lewis and Marafioti, Andrés and Kydlíček, Hynek and Lajarín, Agustín Piqueres and Srivastav, Vaibhav and Lochner, Joshua and Fahlgren, Caleb and Nguyen, Xuan-Son and Fourrier, Clémentine and Burtenshaw, Ben and Larcher, Hugo and Zhao, Haojun and Zakka, Cyril and Morlon, Mathieu and Raffel, Colin and Werra, Leandro von and Wolf, Thomas},
	year = {2025},
}

@article{miles_understanding_2024,
	title = {Understanding the {Role} of the {Projector} in {Knowledge} {Distillation}},
	volume = {38},
	copyright = {Copyright (c) 2024 Association for the Advancement of Artificial Intelligence},
	issn = {2374-3468},
	url = {https://ojs.aaai.org/index.php/AAAI/article/view/28219},
	doi = {10.1609/aaai.v38i5.28219},
	language = {en},
	number = {5},
	urldate = {2026-09-22},
	journal = {Proceedings of the AAAI Conference on Artificial Intelligence},
	author = {Miles, Roy and Mikolajczyk, Krystian},
	month = mar,
	year = {2024},
	pages = {4233--4241},
}

@inproceedings{chen_improved_2022,
	title = {Improved {Feature} {Distillation} via {Projector} {Ensemble}},
	volume = {35},
	url = {https://proceedings.neurips.cc/paper_files/paper/2022/file/4ec0b6648bdf487a2f1c815924339022-Paper-Conference.pdf},
	doi = {10.52202/068431-0878},
	booktitle = {Advances in {Neural} {Information} {Processing} {Systems}},
	publisher = {Curran Associates, Inc.},
	author = {Chen, Yudong and Wang, Sen and Liu, Jiajun and Xu, Xuwei and de Hoog, Frank and Huang, Zi},
	editor = {Koyejo, S. and Mohamed, S. and Agarwal, A. and Belgrave, D. and Cho, K. and Oh, A.},
	year = {2022},
	pages = {12084--12095},
}

@misc{lakkapragada_mitigating_2022,
	title = {Mitigating {Negative} {Transfer} in {Multi}-{Task} {Learning} with {Exponential} {Moving} {Average} {Loss} {Weighting} {Strategies}},
	url = {https://arxiv.org/abs/2211.12999},
	author = {Lakkapragada, Anish and Sleiman, Essam and Surabhi, Saimourya and Wall, Dennis P.},
	year = {2022},
}

@misc{kendall_multi-task_2018,
	title = {Multi-{Task} {Learning} {Using} {Uncertainty} to {Weigh} {Losses} for {Scene} {Geometry} and {Semantics}},
	url = {https://arxiv.org/abs/1705.07115},
	author = {Kendall, Alex and Gal, Yarin and Cipolla, Roberto},
	year = {2018},
}

@misc{li_dynamic_2021,
	title = {Dynamic {Knowledge} {Distillation} for {Pre}-trained {Language} {Models}},
	url = {https://arxiv.org/abs/2109.11295},
	author = {Li, Lei and Lin, Yankai and Ren, Shuhuai and Li, Peng and Zhou, Jie and Sun, Xu},
	year = {2021},
}

@misc{wang_improved_2023,
	title = {Improved {Knowledge} {Distillation} for {Pre}-trained {Language} {Models} via {Knowledge} {Selection}},
	url = {https://arxiv.org/abs/2302.00444},
	author = {Wang, Chenglong and Lu, Yi and Mu, Yongyu and Hu, Yimin and Xiao, Tong and Zhu, Jingbo},
	year = {2023},
}

@misc{park_geometry_2025,
	title = {The {Geometry} of {Categorical} and {Hierarchical} {Concepts} in {Large} {Language} {Models}},
	url = {https://arxiv.org/abs/2406.01506},
	author = {Park, Kiho and Choe, Yo Joong and Jiang, Yibo and Veitch, Victor},
	year = {2025},
}

@misc{gurnee_language_2024,
	title = {Language {Models} {Represent} {Space} and {Time}},
	url = {https://arxiv.org/abs/2310.02207},
	author = {Gurnee, Wes and Tegmark, Max},
	year = {2024},
}

@misc{rein_gpqa_2023,
	title = {{GPQA}: {A} {Graduate}-{Level} {Google}-{Proof} {Q}\&{A} {Benchmark}},
	url = {https://arxiv.org/abs/2311.12022},
	author = {Rein, David and Hou, Betty Li and Stickland, Asa Cooper and Petty, Jackson and Pang, Richard Yuanzhe and Dirani, Julien and Michael, Julian and Bowman, Samuel R.},
	year = {2023},
}

@misc{suzgun_challenging_2022,
	title = {Challenging {BIG}-{Bench} {Tasks} and {Whether} {Chain}-of-{Thought} {Can} {Solve} {Them}},
	url = {https://arxiv.org/abs/2210.09261},
	author = {Suzgun, Mirac and Scales, Nathan and Schärli, Nathanael and Gehrmann, Sebastian and Tay, Yi and Chung, Hyung Won and Chowdhery, Aakanksha and Le, Quoc V. and Chi, Ed H. and Zhou, Denny and Wei, Jason},
	year = {2022},
}

@misc{qwen_qwen25_2025,
	title = {Qwen2.5 {Technical} {Report}},
	url = {https://arxiv.org/abs/2412.15115},
	author = {{Qwen} and {:} and Yang, An and Yang, Baosong and Zhang, Beichen and Hui, Binyuan and Zheng, Bo and Yu, Bowen and Li, Chengyuan and Liu, Dayiheng and Huang, Fei and Wei, Haoran and Lin, Huan and Yang, Jian and Tu, Jianhong and Zhang, Jianwei and Yang, Jianxin and Yang, Jiaxi and Zhou, Jingren and Lin, Junyang and Dang, Kai and Lu, Keming and Bao, Keqin and Yang, Kexin and Yu, Le and Li, Mei and Xue, Mingfeng and Zhang, Pei and Zhu, Qin and Men, Rui and Lin, Runji and Li, Tianhao and Tang, Tianyi and Xia, Tingyu and Ren, Xingzhang and Ren, Xuancheng and Fan, Yang and Su, Yang and Zhang, Yichang and Wan, Yu and Liu, Yuqiong and Cui, Zeyu and Zhang, Zhenru and Qiu, Zihan},
	year = {2025},
}

@misc{guha_openthoughts_2025,
	title = {{OpenThoughts}: {Data} {Recipes} for {Reasoning} {Models}},
	url = {https://arxiv.org/abs/2506.04178},
	author = {Guha, Etash and Marten, Ryan and Keh, Sedrick and Raoof, Negin and Smyrnis, Georgios and Bansal, Hritik and Nezhurina, Marianna and Mercat, Jean and Vu, Trung and Sprague, Zayne and Suvarna, Ashima and Feuer, Benjamin and Chen, Liangyu and Khan, Zaid and Frankel, Eric and Grover, Sachin and Choi, Caroline and Muennighoff, Niklas and Su, Shiye and Zhao, Wanjia and Yang, John and Pimpalgaonkar, Shreyas and Sharma, Kartik and Ji, Charlie Cheng-Jie and Deng, Yichuan and Pratt, Sarah and Ramanujan, Vivek and Saad-Falcon, Jon and Li, Jeffrey and Dave, Achal and Albalak, Alon and Arora, Kushal and Wulfe, Blake and Hegde, Chinmay and Durrett, Greg and Oh, Sewoong and Bansal, Mohit and Gabriel, Saadia and Grover, Aditya and Chang, Kai-Wei and Shankar, Vaishaal and Gokaslan, Aaron and Merrill, Mike A. and Hashimoto, Tatsunori and Choi, Yejin and Jitsev, Jenia and Heckel, Reinhard and Sathiamoorthy, Maheswaran and Dimakis, Alexandros G. and Schmidt, Ludwig},
	year = {2025},
}

@misc{martinsson_randomized_2021,
	title = {Randomized {Numerical} {Linear} {Algebra}: {Foundations} \& {Algorithms}},
	url = {https://arxiv.org/abs/2002.01387},
	author = {Martinsson, Per-Gunnar and Tropp, Joel},
	year = {2021},
}

@misc{mahoney_randomized_2011,
	title = {Randomized algorithms for matrices and data},
	url = {https://arxiv.org/abs/1104.5557},
	author = {Mahoney, Michael W.},
	year = {2011},
}

@misc{gudibande_false_2023,
	title = {The {False} {Promise} of {Imitating} {Proprietary} {LLMs}},
	url = {https://arxiv.org/abs/2305.15717},
	author = {Gudibande, Arnav and Wallace, Eric and Snell, Charlie and Geng, Xinyang and Liu, Hao and Abbeel, Pieter and Levine, Sergey and Song, Dawn},
	year = {2023},
}

@article{guo_deepseek-r1_2025,
	title = {{DeepSeek}-{R1} incentivizes reasoning in {LLMs} through reinforcement learning},
	volume = {645},
	issn = {1476-4687},
	url = {http://dx.doi.org/10.1038/s41586-025-09422-z},
	doi = {10.1038/s41586-025-09422-z},
	number = {8081},
	journal = {Nature},
	publisher = {Springer Science and Business Media LLC},
	author = {Guo, Daya and Yang, Dejian and Zhang, Haowei and Song, Junxiao and Wang, Peiyi and Zhu, Qihao and Xu, Runxin and Zhang, Ruoyu and Ma, Shirong and Bi, Xiao and Zhang, Xiaokang and Yu, Xingkai and Wu, Yu and Wu, Z. F. and Gou, Zhibin and Shao, Zhihong and Li, Zhuoshu and Gao, Ziyi and Liu, Aixin and Xue, Bing and Wang, Bingxuan and Wu, Bochao and Feng, Bei and Lu, Chengda and Zhao, Chenggang and Deng, Chengqi and Ruan, Chong and Dai, Damai and Chen, Deli and Ji, Dongjie and Li, Erhang and Lin, Fangyun and Dai, Fucong and Luo, Fuli and Hao, Guangbo and Chen, Guanting and Li, Guowei and Zhang, H. and Xu, Hanwei and Ding, Honghui and Gao, Huazuo and Qu, Hui and Li, Hui and Guo, Jianzhong and Li, Jiashi and Chen, Jingchang and Yuan, Jingyang and Tu, Jinhao and Qiu, Junjie and Li, Junlong and Cai, J. L. and Ni, Jiaqi and Liang, Jian and Chen, Jin and Dong, Kai and Hu, Kai and You, Kaichao and Gao, Kaige and Guan, Kang and Huang, Kexin and Yu, Kuai and Wang, Lean and Zhang, Lecong and Zhao, Liang and Wang, Litong and Zhang, Liyue and Xu, Lei and Xia, Leyi and Zhang, Mingchuan and Zhang, Minghua and Tang, Minghui and Zhou, Mingxu and Li, Meng and Wang, Miaojun and Li, Mingming and Tian, Ning and Huang, Panpan and Zhang, Peng and Wang, Qiancheng and Chen, Qinyu and Du, Qiushi and Ge, Ruiqi and Zhang, Ruisong and Pan, Ruizhe and Wang, Runji and Chen, R. J. and Jin, R. L. and Chen, Ruyi and Lu, Shanghao and Zhou, Shangyan and Chen, Shanhuang and Ye, Shengfeng and Wang, Shiyu and Yu, Shuiping and Zhou, Shunfeng and Pan, Shuting and Li, S. S. and Zhou, Shuang and Wu, Shaoqing and Yun, Tao and Pei, Tian and Sun, Tianyu and Wang, T. and Zeng, Wangding and Liu, Wen and Liang, Wenfeng and Gao, Wenjun and Yu, Wenqin and Zhang, Wentao and Xiao, W. L. and An, Wei and Liu, Xiaodong and Wang, Xiaohan and Chen, Xiaokang and Nie, Xiaotao and Cheng, Xin and Liu, Xin and Xie, Xin and Liu, Xingchao and Yang, Xinyu and Li, Xinyuan and Su, Xuecheng and Lin, Xuheng and Li, X. Q. and Jin, Xiangyue and Shen, Xiaojin and Chen, Xiaosha and Sun, Xiaowen and Wang, Xiaoxiang and Song, Xinnan and Zhou, Xinyi and Wang, Xianzu and Shan, Xinxia and Li, Y. K. and Wang, Y. Q. and Wei, Y. X. and Zhang, Yang and Xu, Yanhong and Li, Yao and Zhao, Yao and Sun, Yaofeng and Wang, Yaohui and Yu, Yi and Zhang, Yichao and Shi, Yifan and Xiong, Yiliang and He, Ying and Piao, Yishi and Wang, Yisong and Tan, Yixuan and Ma, Yiyang and Liu, Yiyuan and Guo, Yongqiang and Ou, Yuan and Wang, Yuduan and Gong, Yue and Zou, Yuheng and He, Yujia and Xiong, Yunfan and Luo, Yuxiang and You, Yuxiang and Liu, Yuxuan and Zhou, Yuyang and Zhu, Y. X. and Huang, Yanping and Li, Yaohui and Zheng, Yi and Zhu, Yuchen and Ma, Yunxian and Tang, Ying and Zha, Yukun and Yan, Yuting and Ren, Z. Z. and Ren, Zehui and Sha, Zhangli and Fu, Zhe and Xu, Zhean and Xie, Zhenda and Zhang, Zhengyan and Hao, Zhewen and Ma, Zhicheng and Yan, Zhigang and Wu, Zhiyu and Gu, Zihui and Zhu, Zijia and Liu, Zijun and Li, Zilin and Xie, Ziwei and Song, Ziyang and Pan, Zizheng and Huang, Zhen and Xu, Zhipeng and Zhang, Zhongyu and Zhang, Zhen},
	month = sep,
	year = {2025},
	pages = {633--638},
}

@misc{meta_llama_2024,
	title = {Llama 3.2: {Revolutionizing} edge {AI} and vision with open, customizable models},
	url = {https://ai.meta.com/blog/llama-3-2-connect-2024-vision-edge-mobile-devices/},
	author = {{Meta}},
	month = sep,
	year = {2024},
}

@misc{team_gemma_2024,
	title = {Gemma 2: {Improving} {Open} {Language} {Models} at a {Practical} {Size}},
	url = {https://arxiv.org/abs/2408.00118},
	author = {Team, Gemma and Riviere, Morgane and Pathak, Shreya and Sessa, Pier Giuseppe and Hardin, Cassidy and Bhupatiraju, Surya and Hussenot, Léonard and Mesnard, Thomas and Shahriari, Bobak and Ramé, Alexandre and Ferret, Johan and Liu, Peter and Tafti, Pouya and Friesen, Abe and Casbon, Michelle and Ramos, Sabela and Kumar, Ravin and Lan, Charline Le and Jerome, Sammy and Tsitsulin, Anton and Vieillard, Nino and Stanczyk, Piotr and Girgin, Sertan and Momchev, Nikola and Hoffman, Matt and Thakoor, Shantanu and Grill, Jean-Bastien and Neyshabur, Behnam and Bachem, Olivier and Walton, Alanna and Severyn, Aliaksei and Parrish, Alicia and Ahmad, Aliya and Hutchison, Allen and Abdagic, Alvin and Carl, Amanda and Shen, Amy and Brock, Andy and Coenen, Andy and Laforge, Anthony and Paterson, Antonia and Bastian, Ben and Piot, Bilal and Wu, Bo and Royal, Brandon and Chen, Charlie and Kumar, Chintu and Perry, Chris and Welty, Chris and Choquette-Choo, Christopher A. and Sinopalnikov, Danila and Weinberger, David and Vijaykumar, Dimple and Rogozińska, Dominika and Herbison, Dustin and Bandy, Elisa and Wang, Emma and Noland, Eric and Moreira, Erica and Senter, Evan and Eltyshev, Evgenii and Visin, Francesco and Rasskin, Gabriel and Wei, Gary and Cameron, Glenn and Martins, Gus and Hashemi, Hadi and Klimczak-Plucińska, Hanna and Batra, Harleen and Dhand, Harsh and Nardini, Ivan and Mein, Jacinda and Zhou, Jack and Svensson, James and Stanway, Jeff and Chan, Jetha and Zhou, Jin Peng and Carrasqueira, Joana and Iljazi, Joana and Becker, Jocelyn and Fernandez, Joe and Amersfoort, Joost van and Gordon, Josh and Lipschultz, Josh and Newlan, Josh and Ji, Ju-yeong and Mohamed, Kareem and Badola, Kartikeya and Black, Kat and Millican, Katie and McDonell, Keelin and Nguyen, Kelvin and Sodhia, Kiranbir and Greene, Kish and Sjoesund, Lars Lowe and Usui, Lauren and Sifre, Laurent and Heuermann, Lena and Lago, Leticia and McNealus, Lilly and Soares, Livio Baldini and Kilpatrick, Logan and Dixon, Lucas and Martins, Luciano and Reid, Machel and Singh, Manvinder and Iverson, Mark and Görner, Martin and Velloso, Mat and Wirth, Mateo and Davidow, Matt and Miller, Matt and Rahtz, Matthew and Watson, Matthew and Risdal, Meg and Kazemi, Mehran and Moynihan, Michael and Zhang, Ming and Kahng, Minsuk and Park, Minwoo and Rahman, Mofi and Khatwani, Mohit and Dao, Natalie and Bardoliwalla, Nenshad and Devanathan, Nesh and Dumai, Neta and Chauhan, Nilay and Wahltinez, Oscar and Botarda, Pankil and Barnes, Parker and Barham, Paul and Michel, Paul and Jin, Pengchong and Georgiev, Petko and Culliton, Phil and Kuppala, Pradeep and Comanescu, Ramona and Merhej, Ramona and Jana, Reena and Rokni, Reza Ardeshir and Agarwal, Rishabh and Mullins, Ryan and Saadat, Samaneh and Carthy, Sara Mc and Cogan, Sarah and Perrin, Sarah and Arnold, Sébastien M. R. and Krause, Sebastian and Dai, Shengyang and Garg, Shruti and Sheth, Shruti and Ronstrom, Sue and Chan, Susan and Jordan, Timothy and Yu, Ting and Eccles, Tom and Hennigan, Tom and Kocisky, Tomas and Doshi, Tulsee and Jain, Vihan and Yadav, Vikas and Meshram, Vilobh and Dharmadhikari, Vishal and Barkley, Warren and Wei, Wei and Ye, Wenming and Han, Woohyun and Kwon, Woosuk and Xu, Xiang and Shen, Zhe and Gong, Zhitao and Wei, Zichuan and Cotruta, Victor and Kirk, Phoebe and Rao, Anand and Giang, Minh and Peran, Ludovic and Warkentin, Tris and Collins, Eli and Barral, Joelle and Ghahramani, Zoubin and Hadsell, Raia and Sculley, D. and Banks, Jeanine and Dragan, Anca and Petrov, Slav and Vinyals, Oriol and Dean, Jeff and Hassabis, Demis and Kavukcuoglu, Koray and Farabet, Clement and Buchatskaya, Elena and Borgeaud, Sebastian and Fiedel, Noah and Joulin, Armand and Kenealy, Kathleen and Dadashi, Robert and Andreev, Alek},
	year = {2024},
}

@misc{glm-5-team_glm-5_2026,
	title = {{GLM}-5: from {Vibe} {Coding} to {Agentic} {Engineering}},
	url = {https://arxiv.org/abs/2602.15763},
	author = {{GLM-5-Team} and {:} and Zeng, Aohan and Lv, Xin and Hou, Zhenyu and Du, Zhengxiao and Zheng, Qinkai and Chen, Bin and Yin, Da and Ge, Chendi and Huang, Chenghua and Xie, Chengxing and Zhu, Chenzheng and Yin, Congfeng and Wang, Cunxiang and Pan, Gengzheng and Zeng, Hao and Zhang, Haoke and Wang, Haoran and Chen, Huilong and Zhang, Jiajie and Jiao, Jian and Guo, Jiaqi and Wang, Jingsen and Du, Jingzhao and Wu, Jinzhu and Wang, Kedong and Li, Lei and Fan, Lin and Zhong, Lucen and Liu, Mingdao and Zhao, Mingming and Du, Pengfan and Dong, Qian and Lu, Rui and {Shuang-Li} and Cao, Shulin and Liu, Song and Jiang, Ting and Chen, Xiaodong and Zhang, Xiaohan and Huang, Xuancheng and Dong, Xuezhen and Xu, Yabo and Wei, Yao and An, Yifan and Niu, Yilin and Zhu, Yitong and Wen, Yuanhao and Cen, Yukuo and Bai, Yushi and Qiao, Zhongpei and Wang, Zihan and Wang, Zikang and Zhu, Zilin and Liu, Ziqiang and Li, Zixuan and Wang, Bojie and Wen, Bosi and Huang, Can and Cai, Changpeng and Yu, Chao and Li, Chen and Hu, Chengwei and Zhang, Chenhui and Zhang, Dan and Lin, Daoyan and Yang, Dayong and Wang, Di and Ai, Ding and Zhu, Erle and Yi, Fangzhou and Chen, Feiyu and Wen, Guohong and Sun, Hailong and Zhao, Haisha and Hu, Haiyi and Zhang, Hanchen and Liu, Hanrui and Zhang, Hanyu and Peng, Hao and Tai, Hao and Zhang, Haobo and Liu, He and Wang, Hongwei and Yan, Hongxi and Ge, Hongyu and Liu, Huan and Chu, Huanpeng and Zhao, Jia'ni and Wang, Jiachen and Zhao, Jiajing and Ren, Jiamin and Wang, Jiapeng and Zhang, Jiaxin and Gui, Jiayi and Zhao, Jiayue and Li, Jijie and An, Jing and Li, Jing and Yuan, Jingwei and Du, Jinhua and Liu, Jinxin and Zhi, Junkai and Duan, Junwen and Zhou, Kaiyue and Wei, Kangjian and Wang, Ke and Luo, Keyun and Zhang, Laiqiang and Sha, Leigang and Xu, Liang and Wu, Lindong and Ding, Lintao and Chen, Lu and Li, Minghao and Lin, Nianyi and Ta, Pan and Zou, Qiang and Song, Rongjun and Yang, Ruiqi and Tu, Shangqing and Yang, Shangtong and Wu, Shaoxiang and Zhang, Shengyan and Li, Shijie and Li, Shuang and Fan, Shuyi and Qin, Wei and Tian, Wei and Zhang, Weining and Yu, Wenbo and Liang, Wenjie and Kuang, Xiang and Cheng, Xiangmeng and Li, Xiangyang and Yan, Xiaoquan and Hu, Xiaowei and Ling, Xiaoying and Fan, Xing and Xia, Xingye and Zhang, Xinyuan and Zhang, Xinze and Pan, Xirui and Zou, Xu and Zhang, Xunkai and Liu, Yadi and Wu, Yandong and Li, Yanfu and Wang, Yidong and Zhu, Yifan and Tan, Yijun and Zhou, Yilin and Pan, Yiming and Zhang, Ying and Su, Yinpei and Geng, Yipeng and Yan, Yong and Tan, Yonglin and Bi, Yuean and Shen, Yuhan and Yang, Yuhao and Li, Yujiang and Liu, Yunan and Wang, Yunqing and Li, Yuntao and Wu, Yurong and Zhang, Yutao and Duan, Yuxi and Zhang, Yuxuan and Liu, Zezhen and Jiang, Zhengtao and Yan, Zhenhe and Zhang, Zheyu and Wei, Zhixiang and Chen, Zhuo and Feng, Zhuoer and Yao, Zijun and Chai, Ziwei and Wang, Ziyuan and Zhang, Zuzhou and Xu, Bin and Huang, Minlie and Wang, Hongning and Li, Juanzi and Dong, Yuxiao and Tang, Jie},
	year = {2026},
}

@misc{deepseek-ai_deepseek-v4_2026,
	title = {{DeepSeek}-{V4}: {Towards} {Highly} {Efficient} {Million}-{Token} {Context} {Intelligence}},
	author = {{DeepSeek-AI}},
	year = {2026},
}

@misc{qwen_team_qwen38-max_2026,
	title = {Qwen3.8-{Max}: {A} {New} {Bar} for {Coding} and {Cowork}},
	url = {https://qwen.ai/blog?id=qwen3.8},
	author = {{Qwen Team}},
	month = aug,
	year = {2026},
}

@misc{team_kimi_2026,
	title = {Kimi {K3}: {Open} {Frontier} {Intelligence}},
	url = {https://arxiv.org/abs/2607.24653},
	author = {Team, Kimi and Bai, Tongtong and Bai, Yifan and Bao, Yiping and C, M. and Cai, Jianfeng and Cai, Xinyuan and Cao, Peizhou and Cao, Yuxuan and Chai, Ziwei and Charles, Y. and Che, H. S. and Chen, Guanduo and Chen, Guangyu and Chen, Guanzheng and Chen, Huarong and Chen, Jia and Chen, Jianlong and Chen, Jun and Chen, Kexin and Chen, Peng and Chen, Ruijue and Chen, Wentao and Chen, Xin and Chen, Yang and Chen, Yanru and Chen, Yifei and Chen, Yingjiang and Chen, Yuankun and Chen, Yujie and Chen, Yutian and Chen, Zhirong and Cheng, Dazhi and Cheng, Yean and Cui, Jialei and Cui, Jingbing and Dai, Anqi and Deng, Jiaqi and Ding, Hao and Ding, Rui and Ding, Shaofeng and Dong, Mengfan and Dong, Mengnan and Dong, Yuhao and Dong, Yuxin and Du, Angang and Du, Chenzhuang and Du, Dikang and Du, Jusen and Du, Yulun and Fan, Yu and Feng, Jing and Feng, Qiulin and Feng, Yichen and Fu, Kelin and Fu, Qiang and Gao, Fuxuan and Gao, Hongcheng and Gao, Jingyue and Gao, Tong and Gao, Weijia and Geng, Shangyi and Gong, Jie and Gong, Linhu and Gong, Shengao and Gong, Xiaochen and Gu, Qizheng and Gu, Yicheng and Guan, Shuhao and Guo, Haiqing and Guo, Shiqi and Guo, Xiang and Guo, Zhengyan and Hao, Beixi and Hao, Wenxin and Hao, Xiaoru and He, Dailan and He, Haotian and He, Lehan and He, Qi and He, Weiran and He, Xinran and He, Xinyi and He, Yibo and He, Yunjia and Hong, Chao and Hong, Tiange and Hu, Hao and Hu, Jiaxi and Hu, Ruikun and Hu, Weiming and Hu, Yangyang and Hu, Zhenxing and Hua, Liang and Huang, Jinbin and Huang, Ke and Huang, Ruiyuan and Huang, Siying and Huang, Weixiao and Huang, Yan and Huang, Zhengjie and Huang, Zhiqi and Hui, Yulong and Jia, Chaobo and Jiang, Yutong and Jiang, Zhejun and Jiang, Zuoyou and Jin, Wenyi and Jin, Xinyi and Jing, Yu and Kong, Huanjun and Lai, Guokun and Li, Aidi and Li, Cheng and Li, Chengyuan and Li, Cong and Li, Fang and Li, Guanyu and Li, Haoyang and Li, Jia and Li, Junxiong and Li, Lei and Li, Letian and Li, Lincan and Li, Weihong and Li, Wentao and Li, Xintong and Li, Yang and Li, Yishen and Li, Yiwei and Li, Yuxiao and Li, Zhaowei and Li, Zhaoxi and Li, Zheming and Li, Zhengxiao and Li, Zhiyuan and Lin, Jiawei and Lin, Xiaohan and Lin, Yibo and Lin, Zichao and Lin, Ziyan and Liu, Bill and Liu, Boxiao and Liu, Chuan and Liu, Liang and Liu, Shaowei and Liu, Shudong and Liu, Shuran and Liu, Tianwei and Liu, Weizhou and Liu, Yangyang and Liu, Yanming and Liu, Yibo and Liu, Yipeng and Liu, Zhengying and Liu, Zhiheng and Lu, Enzhe and Lu, Haoyu and Lu, Linqiang and Lu, Tingzhan and Lu, Zhiyuan and Luo, Aotian and Luo, G. and Luo, Junyu and Luo, Yifan and Lyu, B. and Lyu, Wenzhou and Mao, Shaoguang and Mei, Yuan and Men, Xin and Ni, Minqing and Niu, Yixuan and Pan, Siyuan and Peng, Shujun and Qi, Zhangyang and Qin, Ruoyu and Qin, ZeChao and Qin, Zeyu and Qiu, Haiquan and Qiu, Jianxin and Qiu, Jiezhong and Qu, Bowen and Qu, Yuhao and Shang, Zeyu and Shao, Youbo and Shen, Han and Shi, Jincheng and Shi, Juanfeng and Shi, Lidong and Shi, Shengyuan and Siu, Wingchun and Song, Pengwei and Song, Xiaoxi and Su, Jianlin and Su, Yunfeng and Su, Zhaochen and Sui, Lin and Sun, Jingsong and Sun, Junyao and Sun, Shaoning and Sun, Shuzhe and Sun, Tongyu and Sun, Yujun and Tai, Yunpeng and Tang, Chuning and Tang, Heyi and Tang, Sirui and Tang, Zecheng and Tian, Chaoran and Tian, Rongpeng and Tian, Yu and Tu, Wei and Wang, Chensi and Wang, Chuang and Wang, Chunjie and Wang, Dinglu and Wang, Feng and Wang, Hailong and Wang, Haiming and Wang, Hao and Wang, Hao and Wang, Huaqing and Wang, Hui and Wang, Jiayi and Wang, Jinglong and Wang, Jinhong and Wang, Jiuzheng and Wang, Linian and Wang, Shaobo and Wang, Shenzhi and Wang, Shuyi and Wang, Si and Wang, Siyuan and Wang, Tianfu and Wang, Wenjue and Wang, Xingran and Wang, Xinmei and Wang, Xinyuan and Wang, Xusheng and Wang, Yalin and Wang, Yangkun and Wang, Yao and Wang, Yaoyu and Wang, Yejie and Wang, Yiqin and Wang, Yucheng and Wang, Yuzhi and Wang, Zhaoji and Wang, Zhaowei and Wang, Zhengtao and Wang, Zhenhao and Wang, Zhongsheng and Wang, Zifan and Wei, Chu and Wei, Ming and Wei, Shouxin and Wen, Zichen and Wu, Fan and Wu, Haoning and Wu, Rucong and Wu, Wenhao and Wu, Xiaoxue and Wu, Yingcong and Wu, Yongqi and Wu, Yuxin and Wu, Zijian and Xian, Xinglang and Xiang, Chenxuan and Xiang, Yuye and Xiao, Bocheng and Xiao, Chenjun and Xiao, Xin and Xie, Jin and Xie, Xiaotong and Xie, Yifeng and Xie, Zhe and Xing, Bowei and Xiong, Yiming and Xu, Baosheng and Xu, Boyu and Xu, Jiale and Xu, Jianfan and Xu, Jing and Xu, Jinjing and Xu, L. H. and Xu, Qingtao and Xu, Shuyao and Xu, Suting and Xu, Tiantian and Xu, Tianxiang and Xu, Weixin and Xu, Xinran and Xu, Yangchuan and Xu, Ye and Xu, Yueni and Xu, Ziyao and Xue, Haonan and Yan, Junjie and Yan, Yaoyao and Yang, Fan and Yang, Guangyao and Yang, Hao and Yang, Junwei and Yang, Ruoyu and Yang, Wenjie and Yang, Xiaofei and Yang, Xinyu and Yang, Yi and Yang, Yiling and Yang, Ying and Yang, Yuchen and Yang, Zhen and Yang, Zhilin and Yang, Zian and Yang, Zuhao and Yao, Haotian and Ye, Dan and Ye, Haoran and Ye, Wenjie and Ye, Zhanbo and Yin, Bohong and Yin, Haoxiang and Yin, Xietong and Yu, Chengzhen and Yu, Haozhen and Yu, Longhui and Yu, Shengnan and Yu, Shuying and Yu, Tianxiang and Yuan, Enming and Yuan, Mengjie and Yue, Tongtian and Yue, Wei and Yue, Yang and Zha, Dunyuan and Zhan, Haobing and Zhang, B. H. and Zhang, Dehao and Zhang, Fei and Zhang, Hao and Zhang, Haoyuan and Zhang, Huanyu and Zhang, Jiapei and Zhang, Jiaxuan and Zhang, Jin and Zhang, Kaiyi and Zhang, Miaozhen and Zhang, Puqi and Zhang, Qinglei and Zhang, Rong and Zhang, Rui and Zhang, Shaoshuai and Zhang, Shiyi and Zhang, Xiaobin and Zhang, Xiaoyun and Zhang, Y. and Zhang, Yangkun and Zhang, Ye and Zhang, Yichi and Zhang, Yikun and Zhang, Yizhi and Zhang, Yongting and Zhang, Yu and Zhang, Yutao and Zhang, Yutong and Zhang, Zheng and Zhang, Zijing and Zhao, Bin and Zhao, Chenguang and Zhao, Feifan and Zhao, Jinglun and Zhao, Jinxiang and Zhao, Shuai and Zhao, Wenshuo and Zhao, Xiangyu and Zhao, Xuanle and Zhao, Yikai and Zhao, Zijia and Zheng, Haozhi and Zheng, Huabin and Zheng, Ruihan and Zheng, Shaojie and Zheng, Tengyang and Zhong, Haofeng and Zhong, Lei and Zhong, Longguang and Zhou, M. and Zhou, Qiankang and Zhou, Runjie and Zhou, Ruozhang and Zhou, Xinyu and Zhou, Yiqiao and Zhou, Zaida and Zhu, Jinguo and Zhu, Liya and Zhu, Xinhao and Zhu, Yangjunfeng and Zhu, Yuxuan and Zhu, Zhen and Zhuang, Chen and Zhuang, Weiyu and Zu, Xinxing},
	year = {2026},
}

@misc{park_linear_2024,
	title = {The {Linear} {Representation} {Hypothesis} and the {Geometry} of {Large} {Language} {Models}},
	url = {http://arxiv.org/abs/2311.03658},
	doi = {10.48550/arXiv.2311.03658},
	urldate = {2026-05-19},
	publisher = {arXiv},
	author = {Park, Kiho and Choe, Yo Joong and Veitch, Victor},
	month = jul,
	year = {2024},
	note = {arXiv:2311.03658 [cs.CL]},
}

@article{hendrycks_measuring_2021,
	title = {Measuring {Massive} {Multitask} {Language} {Understanding}},
	journal = {Proceedings of the International Conference on Learning Representations (ICLR)},
	author = {Hendrycks, Dan and Burns, Collin and Basart, Steven and Zou, Andy and Mazeika, Mantas and Song, Dawn and Steinhardt, Jacob},
	year = {2021},
}

@misc{team_qwen3_2025,
	title = {Qwen3 {Technical} {Report}},
	url = {https://arxiv.org/abs/2505.09388},
	author = {Team, Qwen},
	year = {2025},
	note = {\_eprint: 2505.09388},
}

@misc{team_gemma_2025,
	title = {Gemma 3 {Technical} {Report}},
	url = {http://arxiv.org/abs/2503.19786},
	doi = {10.48550/arXiv.2503.19786},
	urldate = {2026-03-06},
	publisher = {arXiv},
	author = {Team, Gemma and Kamath, Aishwarya and Ferret, Johan and Pathak, Shreya and Vieillard, Nino and Merhej, Ramona and Perrin, Sarah and Matejovicova, Tatiana and Ramé, Alexandre and Rivière, Morgane and Rouillard, Louis and Mesnard, Thomas and Cideron, Geoffrey and Grill, Jean-bastien and Ramos, Sabela and Yvinec, Edouard and Casbon, Michelle and Pot, Etienne and Penchev, Ivo and Liu, Gaël and Visin, Francesco and Kenealy, Kathleen and Beyer, Lucas and Zhai, Xiaohai and Tsitsulin, Anton and Busa-Fekete, Robert and Feng, Alex and Sachdeva, Noveen and Coleman, Benjamin and Gao, Yi and Mustafa, Basil and Barr, Iain and Parisotto, Emilio and Tian, David and Eyal, Matan and Cherry, Colin and Peter, Jan-Thorsten and Sinopalnikov, Danila and Bhupatiraju, Surya and Agarwal, Rishabh and Kazemi, Mehran and Malkin, Dan and Kumar, Ravin and Vilar, David and Brusilovsky, Idan and Luo, Jiaming and Steiner, Andreas and Friesen, Abe and Sharma, Abhanshu and Sharma, Abheesht and Gilady, Adi Mayrav and Goedeckemeyer, Adrian and Saade, Alaa and Feng, Alex and Kolesnikov, Alexander and Bendebury, Alexei and Abdagic, Alvin and Vadi, Amit and György, András and Pinto, André Susano and Das, Anil and Bapna, Ankur and Miech, Antoine and Yang, Antoine and Paterson, Antonia and Shenoy, Ashish and Chakrabarti, Ayan and Piot, Bilal and Wu, Bo and Shahriari, Bobak and Petrini, Bryce and Chen, Charlie and Lan, Charline Le and Choquette-Choo, Christopher A. and Carey, C. J. and Brick, Cormac and Deutsch, Daniel and Eisenbud, Danielle and Cattle, Dee and Cheng, Derek and Paparas, Dimitris and Sreepathihalli, Divyashree Shivakumar and Reid, Doug and Tran, Dustin and Zelle, Dustin and Noland, Eric and Huizenga, Erwin and Kharitonov, Eugene and Liu, Frederick and Amirkhanyan, Gagik and Cameron, Glenn and Hashemi, Hadi and Klimczak-Plucińska, Hanna and Singh, Harman and Mehta, Harsh and Lehri, Harshal Tushar and Hazimeh, Hussein and Ballantyne, Ian and Szpektor, Idan and Nardini, Ivan and Pouget-Abadie, Jean and Chan, Jetha and Stanton, Joe and Wieting, John and Lai, Jonathan and Orbay, Jordi and Fernandez, Joseph and Newlan, Josh and Ji, Ju-yeong and Singh, Jyotinder and Black, Kat and Yu, Kathy and Hui, Kevin and Vodrahalli, Kiran and Greff, Klaus and Qiu, Linhai and Valentine, Marcella and Coelho, Marina and Ritter, Marvin and Hoffman, Matt and Watson, Matthew and Chaturvedi, Mayank and Moynihan, Michael and Ma, Min and Babar, Nabila and Noy, Natasha and Byrd, Nathan and Roy, Nick and Momchev, Nikola and Chauhan, Nilay and Sachdeva, Noveen and Bunyan, Oskar and Botarda, Pankil and Caron, Paul and Rubenstein, Paul Kishan and Culliton, Phil and Schmid, Philipp and Sessa, Pier Giuseppe and Xu, Pingmei and Stanczyk, Piotr and Tafti, Pouya and Shivanna, Rakesh and Wu, Renjie and Pan, Renke and Rokni, Reza and Willoughby, Rob and Vallu, Rohith and Mullins, Ryan and Jerome, Sammy and Smoot, Sara and Girgin, Sertan and Iqbal, Shariq and Reddy, Shashir and Sheth, Shruti and Põder, Siim and Bhatnagar, Sijal and Panyam, Sindhu Raghuram and Eiger, Sivan and Zhang, Susan and Liu, Tianqi and Yacovone, Trevor and Liechty, Tyler and Kalra, Uday and Evci, Utku and Misra, Vedant and Roseberry, Vincent and Feinberg, Vlad and Kolesnikov, Vlad and Han, Woohyun and Kwon, Woosuk and Chen, Xi and Chow, Yinlam and Zhu, Yuvein and Wei, Zichuan and Egyed, Zoltan and Cotruta, Victor and Giang, Minh and Kirk, Phoebe and Rao, Anand and Black, Kat and Babar, Nabila and Lo, Jessica and Moreira, Erica and Martins, Luiz Gustavo and Sanseviero, Omar and Gonzalez, Lucas and Gleicher, Zach and Warkentin, Tris and Mirrokni, Vahab and Senter, Evan and Collins, Eli and Barral, Joelle and Ghahramani, Zoubin and Hadsell, Raia and Matias, Yossi and Sculley, D. and Petrov, Slav and Fiedel, Noah and Shazeer, Noam and Vinyals, Oriol and Dean, Jeff and Hassabis, Demis and Kavukcuoglu, Koray and Farabet, Clement and Buchatskaya, Elena and Alayrac, Jean-Baptiste and Anil, Rohan and Dmitry and Lepikhin and Borgeaud, Sebastian and Bachem, Olivier and Joulin, Armand and Andreev, Alek and Hardin, Cassidy and Dadashi, Robert and Hussenot, Léonard},
	month = mar,
	year = {2025},
	note = {arXiv:2503.19786 [cs]},
}

@article{beals_extensions_1984,
	title = {Extensions of {Lipschitz} mappings into a {Hilbert} space},
	volume = {26},
	url = {http://www.ams.org/conm/026/},
	doi = {10.1090/conm/026/737400},
	language = {en},
	urldate = {2025-10-28},
	publisher = {American Mathematical Society},
	author = {Johnson, William B. and Lindenstrauss, Joram},
	editor = {Beals, Richard and Beck, Anatole and Bellow, Alexandra and Hajian, Arshag},
	year = {1984},
	doi = {10.1090/conm/026/737400},
	note = {Book Title: Contemporary Mathematics
ISBN: 9780821850305 9780821876114
Place: Providence, Rhode Island},
	pages = {189--206},
}

@misc{sun_patient_2019,
	title = {Patient {Knowledge} {Distillation} for {BERT} {Model} {Compression}},
	url = {https://arxiv.org/abs/1908.09355},
	author = {Sun, Siqi and Cheng, Yu and Gan, Zhe and Liu, Jingjing},
	year = {2019},
	note = {\_eprint: 1908.09355},
}

@inproceedings{drinea_randomized_2004,
	title = {A {Randomized} {Singular} {Value} {Decomposition} {Algorithm} for {Image} {Processing} {Applications}},
	url = {https://api.semanticscholar.org/CorpusID:5727723},
	author = {Drinea, Eleni and Drineas, Petros and Huggins, Patrick S.},
	year = {2004},
}

@inproceedings{bucilua_model_2006,
	address = {New York, NY, USA},
	series = {{KDD} '06},
	title = {Model compression},
	isbn = {978-1-59593-339-3},
	url = {https://dl.acm.org/doi/10.1145/1150402.1150464},
	doi = {10.1145/1150402.1150464},
	urldate = {2025-11-28},
	booktitle = {Proceedings of the 12th {ACM} {SIGKDD} international conference on {Knowledge} discovery and data mining},
	publisher = {Association for Computing Machinery},
	author = {Buciluǎ, Cristian and Caruana, Rich and Niculescu-Mizil, Alexandru},
	month = aug,
	year = {2006},
	pages = {535--541},
}

@misc{jiao_tinybert_2020,
	title = {{TinyBERT}: {Distilling} {BERT} for {Natural} {Language} {Understanding}},
	url = {https://arxiv.org/abs/1909.10351},
	author = {Jiao, Xiaoqi and Yin, Yichun and Shang, Lifeng and Jiang, Xin and Chen, Xiao and Li, Linlin and Wang, Fang and Liu, Qun},
	year = {2020},
	note = {\_eprint: 1909.10351},
}

@misc{hinton_distilling_2015,
	title = {Distilling the {Knowledge} in a {Neural} {Network}},
	url = {http://arxiv.org/abs/1503.02531},
	doi = {10.48550/arXiv.1503.02531},
	urldate = {2025-10-28},
	publisher = {arXiv},
	author = {Hinton, Geoffrey and Vinyals, Oriol and Dean, Jeff},
	month = mar,
	year = {2015},
	note = {arXiv:1503.02531 [stat]},
}

@article{achlioptas_database-friendly_2003,
	series = {Special {Issue} on {PODS} 2001},
	title = {Database-friendly random projections: {Johnson}-{Lindenstrauss} with binary coins},
	volume = {66},
	issn = {0022-0000},
	shorttitle = {Database-friendly random projections},
	url = {https://www.sciencedirect.com/science/article/pii/S0022000003000254},
	doi = {10.1016/S0022-0000(03)00025-4},
	number = {4},
	urldate = {2025-10-28},
	journal = {Journal of Computer and System Sciences},
	author = {Achlioptas, Dimitris},
	month = jun,
	year = {2003},
	pages = {671--687},
}

@misc{sanh_distilbert_2020,
	title = {{DistilBERT}, a distilled version of {BERT}: smaller, faster, cheaper and lighter},
	shorttitle = {{DistilBERT}, a distilled version of {BERT}},
	url = {http://arxiv.org/abs/1910.01108},
	doi = {10.48550/arXiv.1910.01108},
	urldate = {2025-10-13},
	publisher = {arXiv},
	author = {Sanh, Victor and Debut, Lysandre and Chaumond, Julien and Wolf, Thomas},
	month = mar,
	year = {2020},
	note = {arXiv:1910.01108 [cs]},
}

@misc{xu_survey_2024,
	title = {A {Survey} on {Knowledge} {Distillation} of {Large} {Language} {Models}},
	url = {http://arxiv.org/abs/2402.13116},
	doi = {10.48550/arXiv.2402.13116},
	urldate = {2025-10-13},
	publisher = {arXiv},
	author = {Xu, Xiaohan and Li, Ming and Tao, Chongyang and Shen, Tao and Cheng, Reynold and Li, Jinyang and Xu, Can and Tao, Dacheng and Zhou, Tianyi},
	month = oct,
	year = {2024},
	note = {arXiv:2402.13116 [cs]},
}

@article{gu_minillm_2024,
	title = {{MiniLLM}: {Knowledge} {Distillation} of {Large} {Language} {Models}},
	volume = {2024},
	shorttitle = {{MiniLLM}},
	url = {https://proceedings.iclr.cc/paper_files/paper/2024/hash/8ac015d409635f196f9e3e9dcfb9a94e-Abstract-Conference.html},
	language = {en},
	urldate = {2025-10-13},
	journal = {International Conference on Representation Learning},
	author = {Gu, Yuxian and Dong, Li and Wei, Furu and Huang, Minlie},
	month = may,
	year = {2024},
	pages = {32694--32717},
}

@article{clark_think_2018,
	title = {Think you have {Solved} {Question} {Answering}? {Try} {ARC}, the {AI2} {Reasoning} {Challenge}},
	journal = {arXiv:1803.05457v1},
	author = {Clark, Peter and Cowhey, Isaac and Etzioni, Oren and Khot, Tushar and Sabharwal, Ashish and Schoenick, Carissa and Tafjord, Oyvind},
	year = {2018},
}

@misc{elhage_toy_2022,
	title = {Toy {Models} of {Superposition}},
	url = {http://arxiv.org/abs/2209.10652},
	doi = {10.48550/arXiv.2209.10652},
	urldate = {2024-11-05},
	publisher = {arXiv},
	author = {Elhage, Nelson and Hume, Tristan and Olsson, Catherine and Schiefer, Nicholas and Henighan, Tom and Kravec, Shauna and Hatfield-Dodds, Zac and Lasenby, Robert and Drain, Dawn and Chen, Carol and Grosse, Roger and McCandlish, Sam and Kaplan, Jared and Amodei, Dario and Wattenberg, Martin and Olah, Christopher},
	month = sep,
	year = {2022},
	note = {arXiv:2209.10652},
}
\bibliographystyle{iclr2027_conference}

\newpage

\appendix

\section{Johnson-Lindenstrauss Lemma}
We provide a restatement of the JL lemma here for convenience.

\begin{lemma}[Johnson--Lindenstrauss Lemma]
\label{jl-lemma}
Let $0 < \varepsilon < 1$ and let
$X = \{x_1,\ldots,x_n\} \subset \mathbb{R}^d$.
For any target dimension
\[
k \geq C\frac{\log n}{\varepsilon^2},
\]
where $C>0$ is a universal constant, there exists a linear map
$f:\mathbb{R}^d \to \mathbb{R}^k$ such that, for every
$x_i,x_j \in X$,
\[
(1-\varepsilon)\lVert x_i-x_j\rVert_2^2
\leq
\lVert f(x_i)-f(x_j)\rVert_2^2
\leq
(1+\varepsilon)\lVert x_i-x_j\rVert_2^2.
\]
\end{lemma}

\section{Achlioptas Projector Initialization Algorithms}
\label{a:achlioptas-init}

\begin{figure}[h]
\centering

\begin{minipage}[t]{0.48\textwidth}
\begin{algorithm}[H]
\caption{Dense Binary Coin Random Projection}
\label{alg:achlioptas-dense}
\begin{algorithmic}[1]
\Require $\mathbf{X}\in\mathbb{R}^{n\times d}$, target dimension $k$
\Ensure $\mathbf{Z}\in\mathbb{R}^{n\times k}$

\State Initialize $\mathbf{R}\in\mathbb{R}^{d\times k}$
\For{$i=1,\ldots,d$}
    \For{$j=1,\ldots,k$}
        \State Sample $u\sim\operatorname{Uniform}(0,1)$
        \If{$u<1/2$}
            \State $R_{ij}\gets +1$
        \Else
            \State $R_{ij}\gets -1$
        \EndIf
    \EndFor
\EndFor
\State $\mathbf{Z}\gets
       \frac{1}{\sqrt{k}}\mathbf{X}\mathbf{R}$
\State \Return $\mathbf{Z}$
\end{algorithmic}
\end{algorithm}
\end{minipage}
\hfill
\begin{minipage}[t]{0.48\textwidth}
\begin{algorithm}[H]
\caption{Sparse Ternary Random Projection}
\label{alg:achlioptas-sparse}
\begin{algorithmic}[1]
\Require $\mathbf{X}\in\mathbb{R}^{n\times d}$, target dimension $k$
\Ensure $\mathbf{Z}\in\mathbb{R}^{n\times k}$

\State Initialize $\mathbf{Q}\in\{-1,0,+1\}^{d\times k}$
\For{$i=1,\ldots,d$}
    \For{$j=1,\ldots,k$}
        \State Sample $u\sim\operatorname{Uniform}(0,1)$
        \If{$u<1/6$}
            \State $Q_{ij}\gets +1$
        \ElsIf{$u<5/6$}
            \State $Q_{ij}\gets 0$
        \Else
            \State $Q_{ij}\gets -1$
        \EndIf
    \EndFor
\EndFor
\State $\mathbf{Z}\gets
       \sqrt{\frac{3}{k}}\mathbf{X}\mathbf{Q}$
\State \Return $\mathbf{Z}$
\end{algorithmic}
\end{algorithm}
\end{minipage}

\end{figure}

\section{Contextualized JL Lemma Guarantees}
We provide a concrete example of the theoretical guarantees of the JL lemma using Gemma 3 270M as a student model. The representation space is $k = 640$ dimensions, and we train for
\begin{equation}
    n
    = \frac{40\,\text{M tokens}}
           {1024\,\text{tokens per sample}}
    \approx 40{,}000\ \text{samples}.
\end{equation}

For \(n=40{,}000\), \(k=640\), and a desired confidence of \(99\%\), set the failure probability to \(\delta=0.01\). Since Achlioptas \cite{achlioptas_database-friendly_2003} gives confidence \(1-n^{-\beta}\), solve

$$ n^{-\beta}=\delta $$

to obtain

$$ \beta = \frac{\ln(1/\delta)}{\ln n} = \frac{\ln(100)}{\ln(40{,}000)} \approx 0.435. $$

Substituting this value into the Achlioptas bound gives

$$ 640 \geq \frac{(4+2(0.435))\ln(40{,}000)} {\varepsilon^2/2-\varepsilon^3/3}. $$

Equivalently,

$$ \frac{\varepsilon^2}{2}-\frac{\varepsilon^3}{3} \geq \frac{(4+2(0.435))\ln(40{,}000)}{640} \approx 0.08062. $$

Solving numerically gives

$$ \varepsilon\approx0.489. $$

Therefore, with probability at least \(99\%\), all pairwise squared distances satisfy

$$ 0.511\|x_i-x_j\|_2^2 \leq \|f(x_i)-f(x_j)\|_2^2 \leq 1.489\|x_i-x_j\|_2^2. $$

In terms of ordinary Euclidean distances,

$$ 0.715\|x_i-x_j\|_2 \leq \|f(x_i)-f(x_j)\|_2 \leq 1.220\|x_i-x_j\|_2. $$

Thus, at \(99\%\) confidence, projecting 40,000 points into 640 dimensions guarantees that every distance is between approximately \(71.5\%\) and \(122.0\%\) of its original value
for every one of the $\binom{40{,}000}{2} = 799{,}980{,}000$ sequence pairs.

\section{Stop-Gradient Optimization Mechanics in Dynamic Objective}
\label{a:stop_gradient_proof}

In Section~\ref{ss:dynamic-weight}, we note that all operations computing the dynamic objective weights $w_{i, t}$ from raw losses $\mathcal{L}_{i, t}$ and log-variance parameters $s_i$ are evaluated using a stop-gradient operator ($\text{sg}[\cdot]$). This section provides a formal analysis of why gradient detachment is mathematically necessary and proves the optimization failure modes that arise when $w_{i, t}$ remains fully differentiable with respect to $\mathbf{s} = [s_1, \dots, s_K]^\top$.

\subsection{Fully Differentiable Formulation without Stop-Gradient}

Consider a simplified variant of the normalized uncertainty weighting objective where intermediate weights $w_i(\mathbf{s})$ maintain active autograd graph connections:
\begin{equation}
    \mathcal{J}(\mathbf{s}) = \sum_{j=1}^{K} w_j(\mathbf{s}) \mathcal{L}_j + \sum_{j=1}^{K} s_j, \qquad \text{where } w_i(\mathbf{s}) = \frac{\exp(-s_i)}{\sum_{k=1}^K \exp(-s_k)}.
    \label{eq:app_diff_loss}
\end{equation}

Evaluating the partial derivative of $\mathcal{J}(\mathbf{s})$ with respect to an individual log-variance $s_i$ yields:
\begin{equation}
    \frac{\partial \mathcal{J}}{\partial s_i} = \sum_{j=1}^K \frac{\partial w_j}{\partial s_i} \mathcal{L}_j + 1.
\end{equation}
Using the standard Jacobian properties of softmax-style normalizations:
\begin{equation}
    \frac{\partial w_i}{\partial s_i} = -w_i(1 - w_i), \qquad \frac{\partial w_j}{\partial s_i} = w_i w_j \quad (j \neq i).
\end{equation}
Substituting these derivatives back into the total gradient gives:
\begin{align}
    \frac{\partial \mathcal{J}}{\partial s_i} &= -w_i(1 - w_i)\mathcal{L}_i + \sum_{j \neq i} w_i w_j \mathcal{L}_j + 1 \nonumber \\
    &= -w_i \mathcal{L}_i + w_i \sum_{j=1}^K w_j \mathcal{L}_j + 1 \nonumber \\
    &= 1 + w_i \left( \overline{\mathcal{L}}_w - \mathcal{L}_i \right),
    \label{eq:app_gradient_exact}
\end{align}
where $\overline{\mathcal{L}}_w = \sum_{j=1}^K w_j \mathcal{L}_j$ represents the current weighted-average loss across all active objectives.

\subsection{Failure Modes of the Un-Detached System}

Equation~\ref{eq:app_gradient_exact} exposes two severe optimization vulnerabilities when gradient paths through $w_i(\mathbf{s})$ are preserved:

\paragraph{1. Loss-Evasion Shortcut.}
If objective $i$ experiences a temporary loss spike such that $\mathcal{L}_i > \overline{\mathcal{L}}_w$, the term $1 + w_i(\overline{\mathcal{L}}_w - \mathcal{L}_i)$ can become negative. Under standard gradient descent ($\mathbf{s} \leftarrow \mathbf{s} - \gamma \nabla_{\mathbf{s}}\mathcal{J}$), a negative gradient forces $s_i$ to grow ($s_i \to +\infty$). The optimizer discovers that it can minimize the global objective $\mathcal{J}$ faster by inflating $s_i$ (suppressing $w_i$) rather than updating the network backbone parameters to reduce the underlying task error $\mathcal{L}_i$.

\paragraph{2. Scale-Invariance Collapse to Negative Infinity.}
More critically, normalized uncertainty weights are invariant under uniform translations of the parameter vector $\mathbf{s}$. For any scalar constant $c \in \mathbb{R}$:
\begin{equation}
    w_i(\mathbf{s} + c\mathbf{1}) = \frac{\exp[-(s_i + c)]}{\sum_{j=1}^K \exp[-(s_j + c)]} = \frac{\exp(-c)\exp(-s_i)}{\exp(-c)\sum_{j=1}^K \exp(-s_j)} = w_i(\mathbf{s}).
\end{equation}
However, the linear regularization penalty $\sum s_j$ is strictly shift-variant:
\begin{equation}
    \mathcal{J}(\mathbf{s} + c\mathbf{1}) = \sum_{j=1}^{K} w_j(\mathbf{s}) \mathcal{L}_j + \sum_{j=1}^{K} (s_j + c) = \mathcal{J}(\mathbf{s}) + K \cdot c.
\end{equation}
Taking the limit as $c \to -\infty$:
\begin{equation}
    \lim_{c \to -\infty} \mathcal{J}(\mathbf{s} + c\mathbf{1}) = -\infty.
\end{equation}
Without a stop-gradient operator, the optimizer can exploit this scale symmetry by driving all $s_i \to -\infty$ at a uniform rate. The relative task weights $w_i$ remain fixed, but the total loss vanishes to negative infinity, completely collapsing training stability.

\subsection{Guaranteed Stability via Stop-Gradient Execution}

In our implementation, we enforce stop-gradient execution by calculating $w_{i, t}$ inside a non-differentiable step routine. In the computational graph for $\mathcal{L}_{\text{total}}$, the weights $w_{i, t}$ are treated as constant coefficients:
\begin{equation}
    \mathcal{L}_{\text{total}} = \sum_{i \in \mathcal{A}} \text{sg}[w_{i, t}] \mathcal{L}_{i, t} + \sum_{i \in \mathcal{A}} s_i.
\end{equation}
Because $\frac{\partial \text{sg}[w_{i, t}]}{\partial s_j} = 0$, the partial derivative with respect to any active log-variance simplifies strictly to:
\begin{equation}
    \frac{\partial \mathcal{L}_{\text{total}}}{\partial s_i} = 1, \qquad \forall i \in \mathcal{A}.
\end{equation}
This structural disconnection guarantees that:
\begin{enumerate}
    \item The cross-derivatives $\frac{\partial w_j}{\partial s_i}$ are zeroed out, eliminating the scale-invariance drift $c \to -\infty$.
    \item Task loss gradients $\nabla_\theta \mathcal{L}_{i, t}$ update the model parameters $\theta$, while $s_i$ is updated exclusively by its regularizer, isolating it as a stable baseline estimator.
\end{enumerate}

\section{Gradient Analysis of Projector--Student Responsibilities}
\label{app:responsibilities}
Here we derive the optimization behavior underlying the
projector--student separation discussed in
Section~\ref{ss:responsibilities}. Let
\begin{equation}
    \mathbf{a}_s\in\mathbb{R}^{d_s},
    \qquad
    \mathbf{a}_t\in\mathbb{R}^{d_t},
    \qquad
    d_s<d_t,
\end{equation}
denote student and teacher activations for a single sample. For student
parameters $\boldsymbol{\theta}\in\mathbb{R}^{p}$, define
\begin{equation}
    \mathbf{J}_{\boldsymbol{\theta}}
    =
    \frac{\partial\mathbf{a}_s}
         {\partial\boldsymbol{\theta}^{\top}}
    \in\mathbb{R}^{d_s\times p}.
\end{equation}

\subsection{Fixed Downward Projection}
For the fixed downward-projection loss in
Section~\ref{ss:responsibilities}, differentiation with respect to the
student activation gives
\begin{equation}
    \nabla_{\mathbf{a}_s}\mathcal{L}_{\mathrm{down}}
    =
    \frac{2}{d_s}
    \left(
        \mathbf{a}_s
        -
        \mathbf{H}_{\mathrm{fix}}\mathbf{a}_t
    \right).
\end{equation}
Applying the chain rule yields
\begin{equation}
    \nabla_{\boldsymbol{\theta}}\mathcal{L}_{\mathrm{down}}
    =
    \frac{2}{d_s}
    \mathbf{J}_{\boldsymbol{\theta}}^\top
    \left(
        \mathbf{a}_s
        -
        \mathbf{H}_{\mathrm{fix}}\mathbf{a}_t
    \right).
\end{equation}
Since
\begin{equation}
    \Delta\mathbf{H}_{\mathrm{fix}}=\mathbf{0},
\end{equation}
the projected teacher activation is stationary for a fixed input. The
projector influences learning only through the target
$\mathbf{H}_{\mathrm{fix}}\mathbf{a}_t$; it neither adapts to reduce the
loss nor transforms the residual during its pullback through the
student.

\subsection{Learned Upward Projection}
For the learned upward-projection loss, define the residual
\begin{equation}
    \mathbf{r}_{\mathrm{up}}
    =
    \mathbf{H}_{\mathrm{learn}}\mathbf{a}_s-\mathbf{a}_t.
\end{equation}
The projector gradient is
\begin{equation}
    \nabla_{\mathbf{H}_{\mathrm{learn}}}
    \mathcal{L}_{\mathrm{up}}
    =
    \frac{2}{d_t}
    \mathbf{r}_{\mathrm{up}}\mathbf{a}_s^\top.
\end{equation}
Thus, the projector can directly reduce the matching residual. The
student-activation gradient is
\begin{equation}
    \nabla_{\mathbf{a}_s}\mathcal{L}_{\mathrm{up}}
    =
    \frac{2}{d_t}
    \mathbf{H}_{\mathrm{learn}}^\top\mathbf{r}_{\mathrm{up}},
\end{equation}
which gives
\begin{align}
    \nabla_{\boldsymbol{\theta}}\mathcal{L}_{\mathrm{up}}
    &=
    \mathbf{J}_{\boldsymbol{\theta}}^\top
    \nabla_{\mathbf{a}_s}\mathcal{L}_{\mathrm{up}}
    \\
    &=
    \frac{2}{d_t}
    \mathbf{J}_{\boldsymbol{\theta}}^\top
    \mathbf{H}_{\mathrm{learn}}^\top
    \left(
        \mathbf{H}_{\mathrm{learn}}\mathbf{a}_s-\mathbf{a}_t
    \right).
\end{align}

The learned projector therefore participates in optimization in two
ways: it determines the residual and transforms that residual through
$\mathbf{H}_{\mathrm{learn}}^\top$ before it reaches the student
Jacobian. Both effects change as the projector is updated. Consequently,
low matching loss does not imply that the student alone has reproduced
the teacher representation; part of the alignment may reside in the
learned projector. Fixing the downward projection removes this
ambiguity and ensures that only the student adapts during distillation.

\section{Representational Alignment Results}
\label{a:rep-align}

\begin{table*}[h]
\centering
\small
\setlength{\tabcolsep}{5pt}
\begin{tabular}{llcccccc}
\toprule
\textbf{Model Family} &
\textbf{Method} &
\multicolumn{3}{c}{\textbf{Global}} &
\multicolumn{3}{c}{\textbf{Mapped}} \\
\cmidrule(lr){3-5}
\cmidrule(lr){6-8}
& &
\textbf{Mean $\pm$ Std.} &
\textbf{Median} &
\textbf{$P_{25}$} &
\textbf{Mean $\pm$ Std.} &
\textbf{Median} &
\textbf{$P_{25}$} \\
\midrule
\multirow{2}{*}{Gemma 3}
& Learned Projector
& $0.3318 \pm 0.1819$
& 0.2985
& 0.2017
& $\mathbf{0.3791} \pm 0.1298$
& 0.3630
& \textbf{0.2914} \\
& RAPID
& $\mathbf{0.3442} \pm 0.1875$
& \textbf{0.3231}
& \textbf{0.2069}
& $0.3739 \pm 0.1287$
& \textbf{0.3648}
& 0.2868 \\
\midrule
\multirow{2}{*}{Qwen 2.5}
& Learned Projector
& $\mathbf{0.7860} \pm 0.1651$
& \textbf{0.8410}
& \textbf{0.7061}
& $\mathbf{0.8963} \pm 0.0883$
& \textbf{0.9295}
& \textbf{0.8917} \\
& RAPID
& $0.7810 \pm 0.1693$
& 0.8395
& 0.6944
& $0.8935 \pm 0.0921$
& 0.9275
& 0.8874 \\
\bottomrule
\end{tabular}
\caption{CKA representational similarity results for the learned projector
and RAPID distillation methods.}
\label{tab:cka_similarity}
\end{table*}

\begin{table*}[h]
\centering
\small
\setlength{\tabcolsep}{5pt}
\begin{tabular}{llcccccc}
\toprule
\textbf{Model Family} &
\textbf{Method} &
\multicolumn{3}{c}{\textbf{Global}} &
\multicolumn{3}{c}{\textbf{Mapped}} \\
\cmidrule(lr){3-5}
\cmidrule(lr){6-8}
& &
\textbf{Mean $\pm$ Std.} &
\textbf{Median} &
\textbf{$P_{25}$} &
\textbf{Mean $\pm$ Std.} &
\textbf{Median} &
\textbf{$P_{25}$} \\
\midrule
\multirow{2}{*}{Gemma 3}
& Learned Projector
& $0.3815 \pm 0.1856$
& 0.3440
& 0.2551
& $0.4745 \pm 0.1647$
& 0.4212
& 0.3590 \\
& RAPID
& $\mathbf{0.4239} \pm 0.1729$
& \textbf{0.4215}
& \textbf{0.3171}
& $\mathbf{0.5063} \pm 0.1393$
& \textbf{0.4603}
& \textbf{0.4321} \\
\midrule
\multirow{2}{*}{Qwen 2.5}
& Learned Projector
& $\mathbf{0.6926} \pm 0.1903$
& \textbf{0.7400}
& \textbf{0.5839}
& $\mathbf{0.8515} \pm 0.0880$
& \textbf{0.8757}
& \textbf{0.8439} \\
& RAPID
& $0.6866 \pm 0.1939$
& 0.7363
& 0.5771
& $0.8454 \pm 0.0957$
& 0.8715
& 0.8381 \\
\bottomrule
\end{tabular}
\caption{RSA representational similarity results for the learned projector
and RAPID distillation methods.}
\label{tab:rsa_similarity}
\end{table*}

\begin{figure}[t]
    \centering
    \begin{subfigure}[t]{0.48\textwidth}
        \centering
        \includegraphics[width=\linewidth]{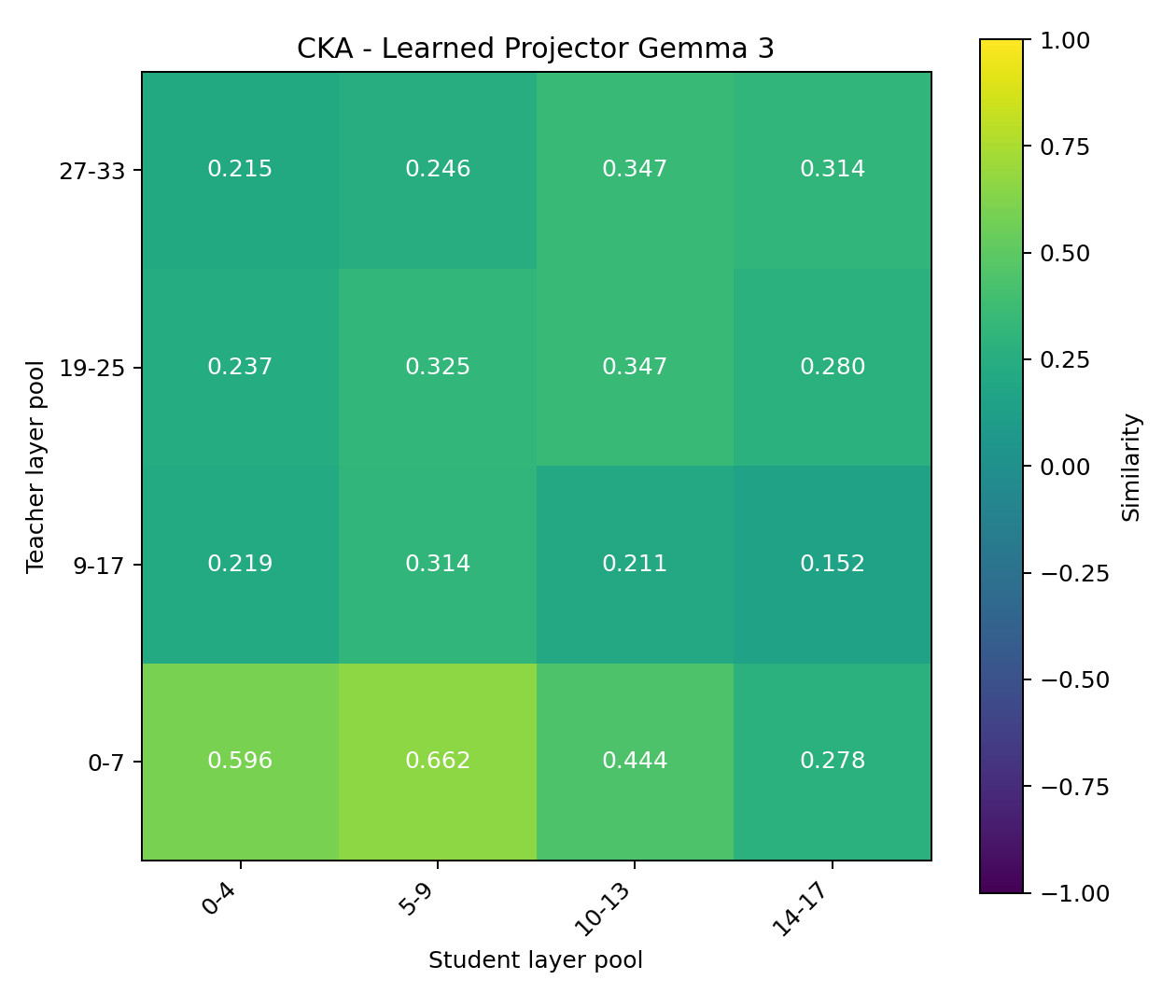}
        \caption{Learned projector}
    \end{subfigure}
    \hfill
    \begin{subfigure}[t]{0.48\textwidth}
        \centering
        \includegraphics[width=\linewidth]{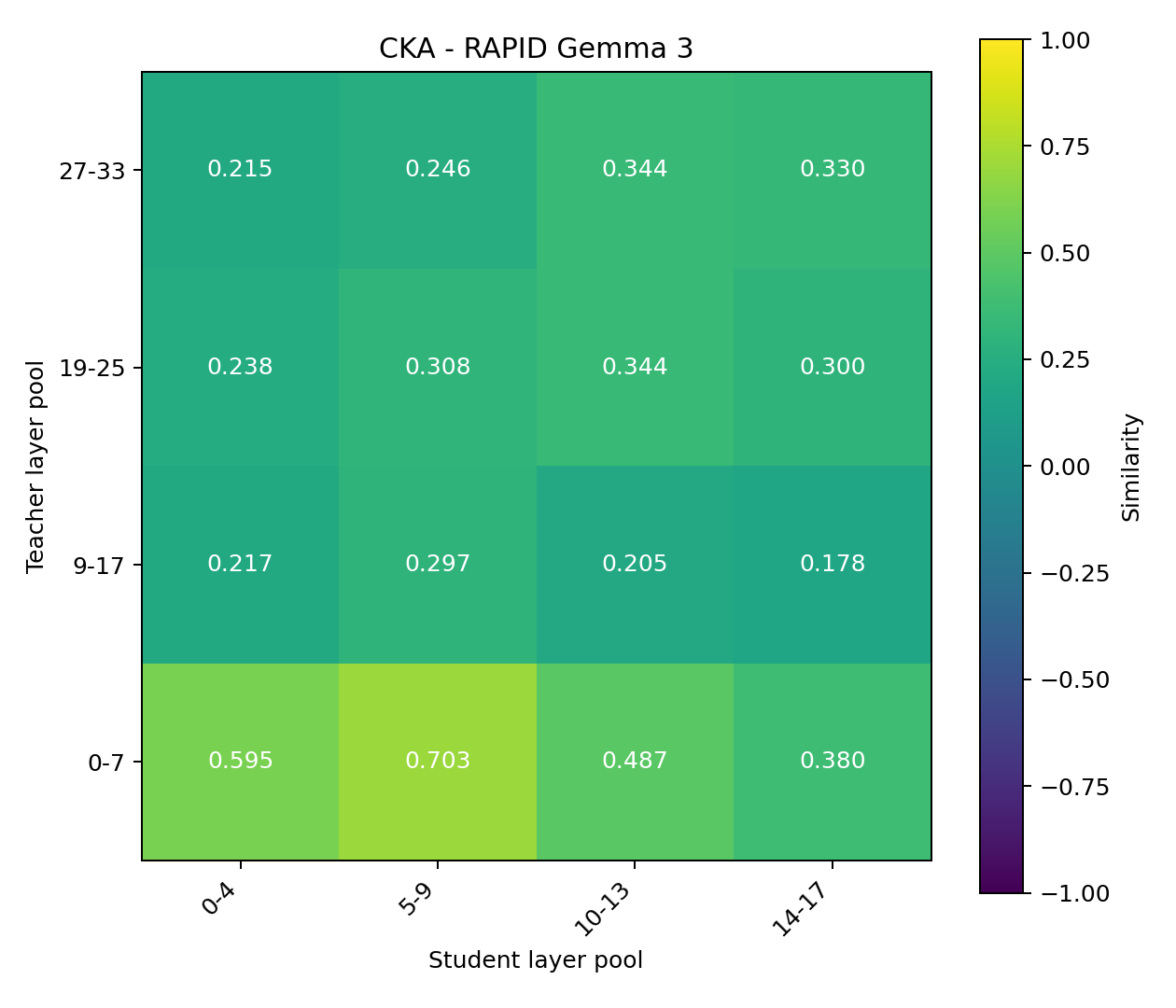}
        \caption{Rapid}
    \end{subfigure}
    \caption{Pooled-layer CKA scores for the Gemma 3 distillation pair comparing RAPID against the traditional learned projector method.}
\end{figure}

\begin{figure}[t]
    \centering
    \begin{subfigure}[t]{0.48\textwidth}
        \centering
        \includegraphics[width=\linewidth]{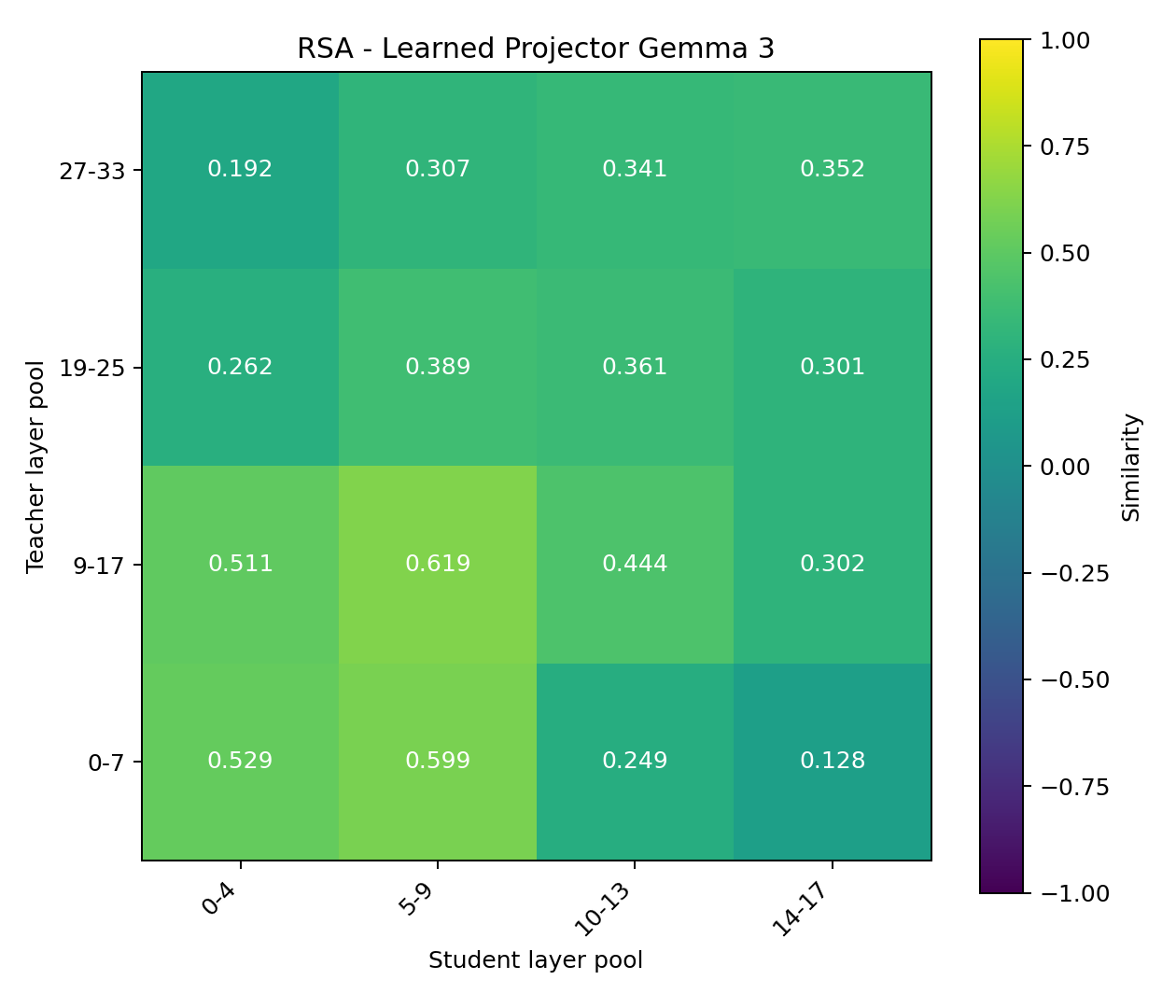}
        \caption{Learned projector}
    \end{subfigure}
    \hfill
    \begin{subfigure}[t]{0.48\textwidth}
        \centering
        \includegraphics[width=\linewidth]{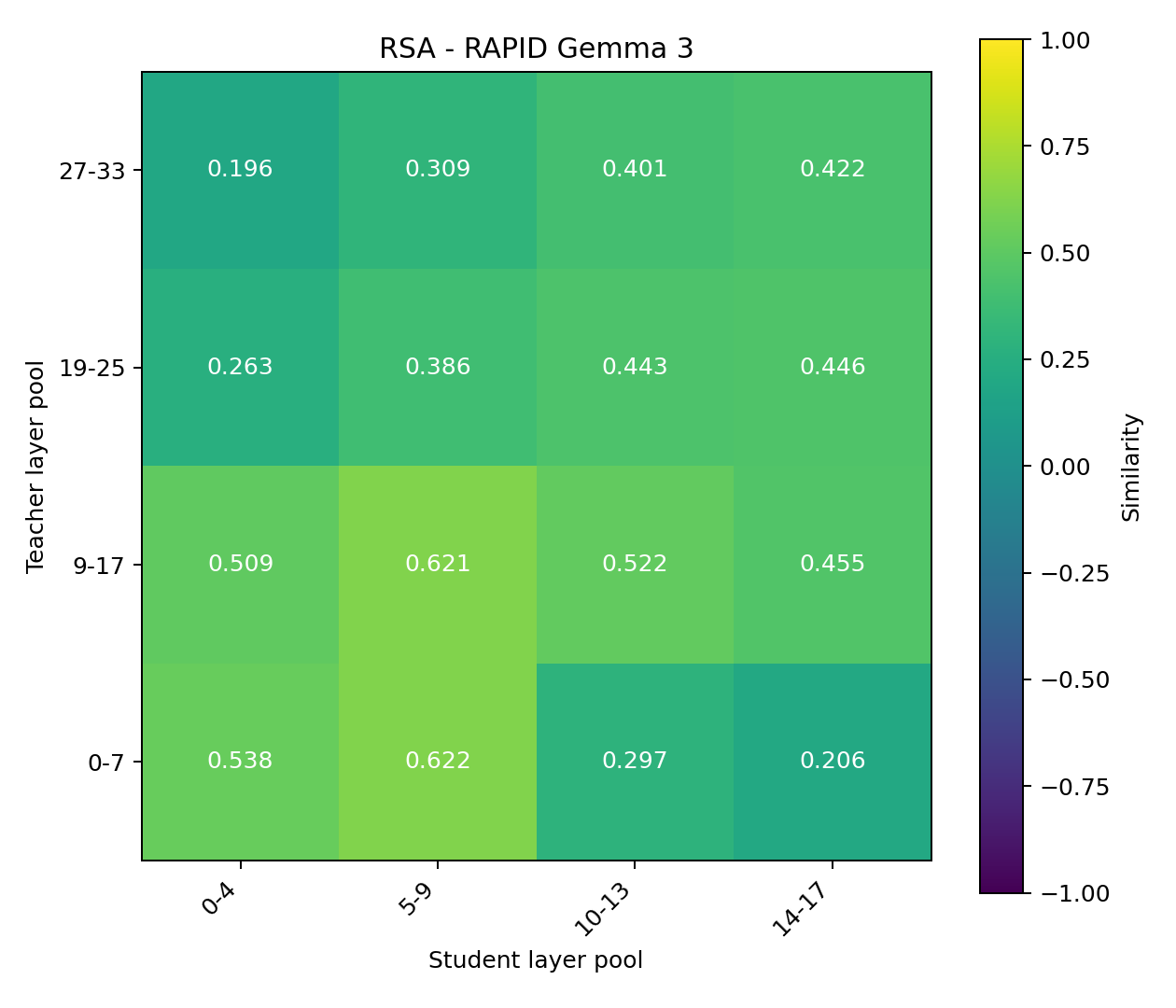}
        \caption{Rapid}
    \end{subfigure}
    \caption{Pooled-layer RSA scores for the Gemma 3 distillation pair comparing RAPID against the traditional learned projector method.}
\end{figure}

\begin{figure}[t]
    \centering
    \begin{subfigure}[t]{0.48\textwidth}
        \centering
        \includegraphics[width=\linewidth]{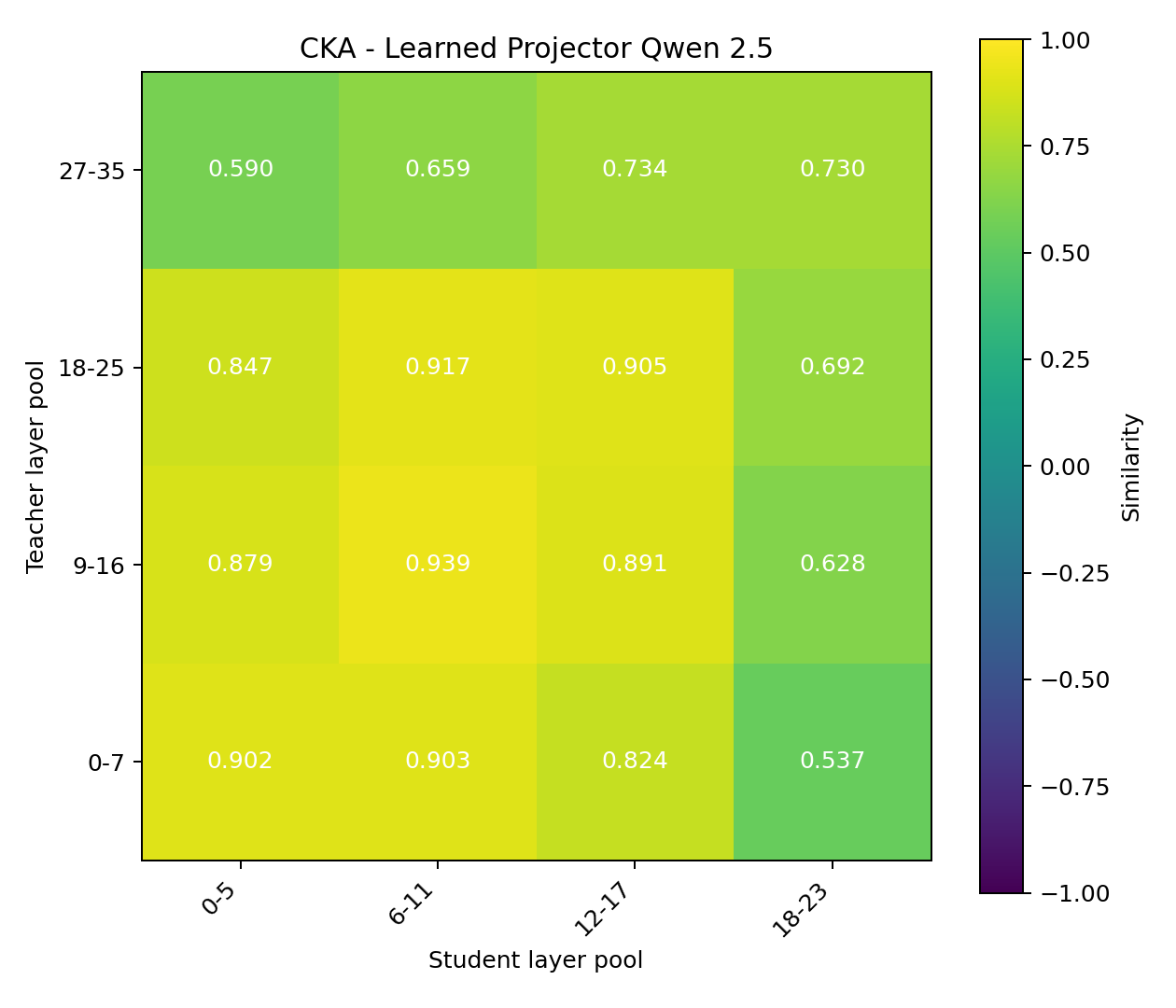}
        \caption{Learned projector}
    \end{subfigure}
    \hfill
    \begin{subfigure}[t]{0.48\textwidth}
        \centering
        \includegraphics[width=\linewidth]{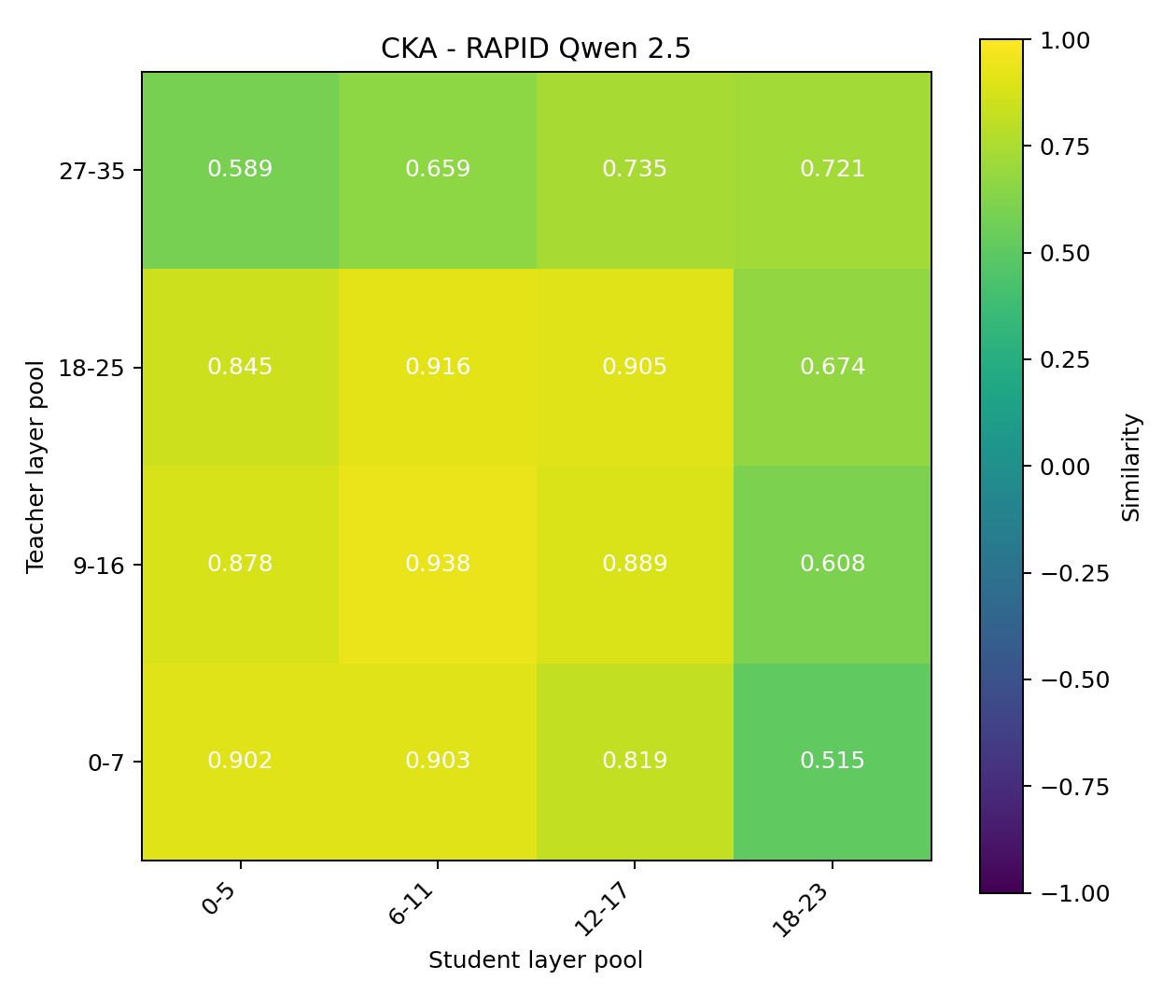}
        \caption{Rapid}
    \end{subfigure}
    \caption{Pooled-layer CKA scores for the Qwen 2.5 distillation pair comparing RAPID against the traditional learned projector method.}
\end{figure}

\begin{figure}[t]
    \centering
    \begin{subfigure}[t]{0.48\textwidth}
        \centering
        \includegraphics[width=\linewidth]{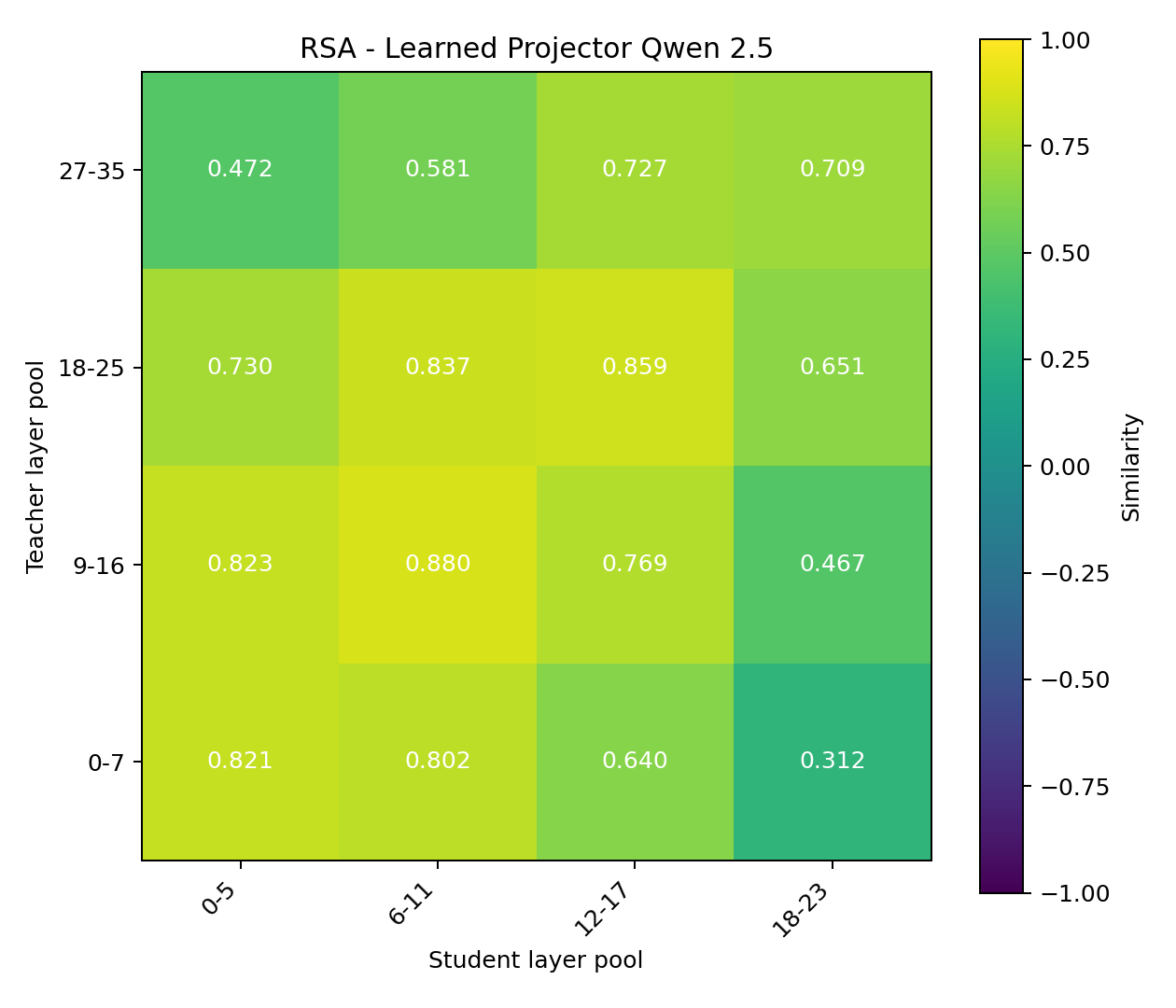}
        \caption{Learned projector}
    \end{subfigure}
    \hfill
    \begin{subfigure}[t]{0.48\textwidth}
        \centering
        \includegraphics[width=\linewidth]{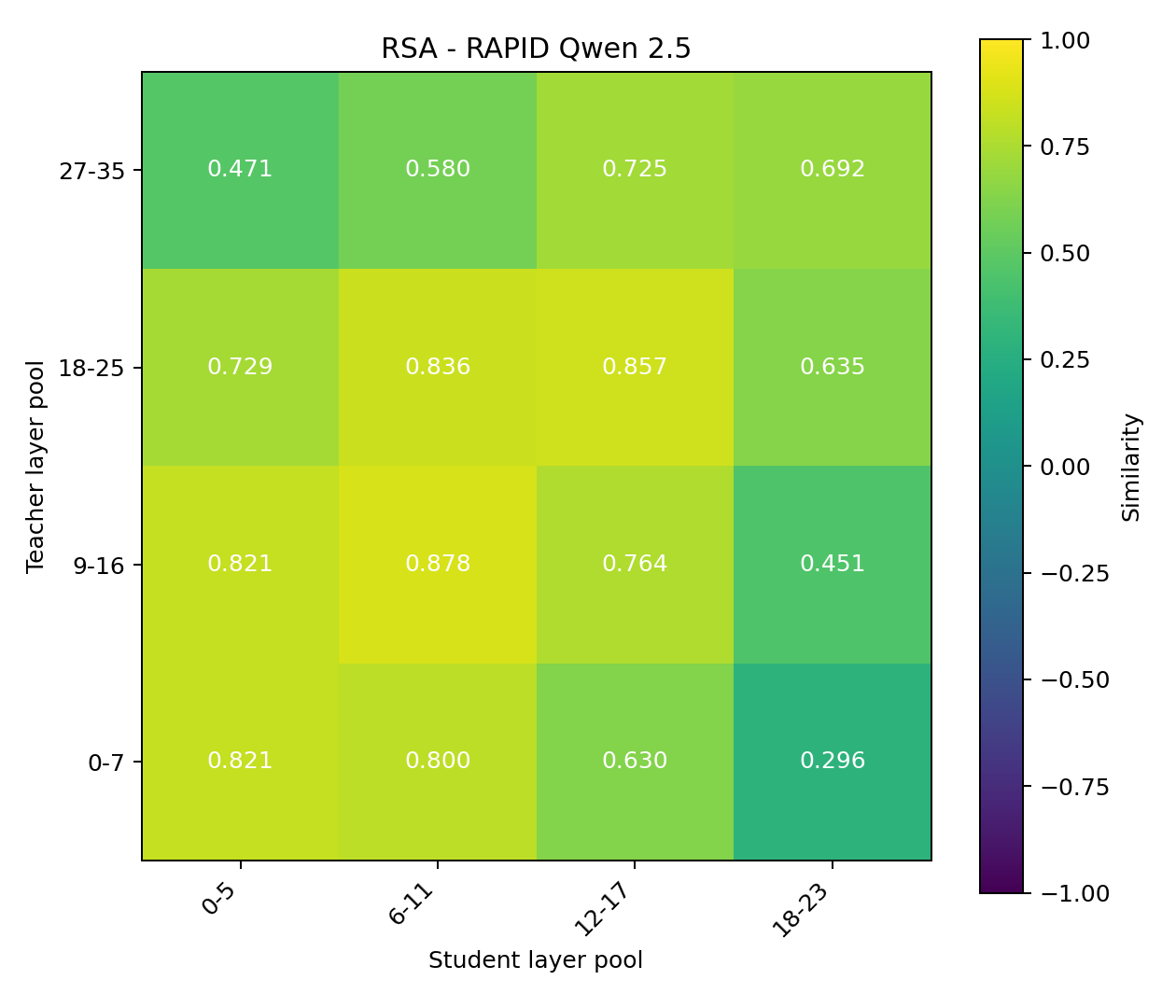}
        \caption{Rapid}
    \end{subfigure}
    \caption{Pooled-layer RSA scores for the Qwen 2.5 distillation pair comparing RAPID against the traditional learned projector method.}
\end{figure}

\clearpage

\section{Training Hyperparameters}

\begin{table}[h]
\centering
\small
\begin{tabular}{lc}
\toprule
\textbf{Hyperparameter} & \textbf{Value} \\
\midrule
Learning rate            & $3 \times 10^{-5}$ \\
Weight decay             & $1 \times 10^{-4}$ \\
Maximum gradient norm    & $2.0$ \\
Distillation temperature & $1.1$ \\
\bottomrule
\end{tabular}
\caption{Training hyperparameters used across all experiments}
\label{tab:hyperparameters}
\end{table}

\end{document}